\documentclass[runningheads]{llncs}

\usepackage{eccv}

\usepackage{eccvabbrv}

\usepackage{graphicx}
\usepackage{booktabs}
\usepackage{comment}
\usepackage{booktabs}
\usepackage{tabularx} 
\usepackage{array} 
\usepackage{ulem}
\usepackage{subcaption}
\usepackage{xcolor}
\usepackage{amssymb}
\usepackage{listings}
\usepackage{dirtree}
\usepackage{tcolorbox}  
\usepackage{multirow}
\usepackage{geometry}   
\usepackage{placeins}
\usepackage{fourier}
\tcbuselibrary{listings}

\usepackage[accsupp]{axessibility}  

\definecolor{eclipseBlue}{RGB}{42,0.0,255}
\definecolor{eclipseGreen}{RGB}{63,127,95}
\definecolor{eclipsePurple}{RGB}{127,0,85}
\definecolor{lightgray}{rgb}{0.95,0.95,0.95}

\lstdefinelanguage{json}{
    basicstyle=\ttfamily\scriptsize,
    numbers=left,
    numberstyle=\tiny\color{gray},
    stepnumber=1,
    numbersep=8pt,
    showstringspaces=false,
    breaklines=true,
    frame=lines,
    backgroundcolor=\color{lightgray},
    stringstyle=\color{eclipseBlue},
    commentstyle=\color{eclipseGreen},
    keywordstyle=\color{eclipsePurple},
    literate=
     *{:}{{{\color{red}{:}}}}{1}
      {,}{{{\color{red}{,}}}}{1}
      {\{}{{{\color{black}\{}}}{1}
      {\}}{{{\color{black}\}}}}{1}
      {[}{{{\color{black}[}}}{1}
      {]}{{{\color{black}]}}}{1},
}

\newtcolorbox{promptbox}[1][]{
  colback=blue!3!white,    
  colframe=blue!60!black,  
  title=#1,
  fonttitle=\bfseries\sffamily,
  fontupper=\small\sffamily, 
  arc=3pt,
  boxrule=0.8pt,
  left=8pt, right=8pt, top=8pt, bottom=8pt
}

\usepackage{hyperref}

\usepackage{orcidlink}

\begin{document}

\title{GroundSet: A Cadastral-Grounded Dataset for Spatial Understanding with Vector Data} 

\titlerunning{GroundSet}

\author{Roger Ferrod\inst{1} \and
Maël Lecene\inst{1} \and
Krishna Sapkota\inst{2} \and
George Leifman\inst{2} \and
Vered Silverman\inst{2} \and \\
Genady Beryozkin\inst{2} \and
Sylvain Lobry\inst{1}
}

\authorrunning{R.Ferrod et al.}

\institute{Université Paris Cité, LIPADE, F-75006 Paris, France \\
\email{sylvain.lobry@u-paris.fr} \and
Google Research \\
}

\maketitle

\begin{abstract}
Precise spatial understanding in Earth Observation is essential for translating raw aerial imagery into actionable insights for critical applications like urban planning, environmental monitoring and disaster management. However, Multimodal Large Language Models exhibit critical deficiencies in fine-grained spatial understanding within Remote Sensing, primarily due to a reliance on limited or repurposed legacy datasets. To bridge this gap, we introduce a large-scale dataset grounded in verifiable cadastral vector data, comprising 3.8 million annotated objects across 510k high-resolution images with 135 granular semantic categories. We validate this resource through a comprehensive instruction-tuning benchmark spanning seven spatial reasoning tasks. Our evaluation establishes a robust baseline using a standard LLaVA architecture. We show that while current RS-specialized and commercial models (e.g., Gemini) struggle in zero-shot settings, high-fidelity supervision effectively bridges this gap, enabling standard architectures to master fine-grained spatial grounding without complex architectural modifications. Data, pretrained model and code are available at: \url{https://huggingface.co/datasets/RogerFerrod/GroundSet}

\end{abstract}

\section{Introduction}
\label{sec:intro}
Multimodal models have become ubiquitous, embedding themselves into any sort of modern application. In particular, the advent of Multimodal Large Language Models (MLLMs) \cite{chatgpt4v, gemini, llava} has fundamentally reshaped the AI landscape. Driven by unprecedented scaling, MLLMs have demonstrated remarkable zero-shot capabilities and broad world knowledge, enabling them to solve a diverse array of tasks with minimal supervision. Furthermore, when integrated into agentic frameworks, these models can orchestrate external tools to address complex, multi-step problems \cite{10.1007/978-3-662-72243-5_13, lahouel2026iceointerpretablecodebasedassistant}. Despite these advancements, however, a critical limitation persists: current MLLMs exhibit a severe deficiency in fine-grained spatial understanding \cite{Bai2024HallucinationOM, adejumo2025visioncentricremotesensing}. They significantly struggle with fundamental vision tasks -- such as precise object detection and segmentation -- which are prerequisites for deployment in high-stakes domains like Remote Sensing (RS), where sub-meter precision is critical and current MLLM inaccuracies actively prevent operational use.

The adaptation of MLLMs to Earth Observation (EO) is compelling yet practically stalled by the scarcity of high-quality spatial annotations. While unlabelled satellite imagery is abundant, dense and accurate annotations remain a bottleneck: human labeling is cost-prohibitive and lacks the scalability required to support modern machine learning research. Consequently, recent literature (e.g., \cite{geochat, skysensegpt, vrsbench, geopixel}) has increasingly relied on repurposing legacy datasets originally designed for closed-set object detection (e.g., DOTA \cite{dota}, DIOR \cite{dior}). These works typically transform older data by synthetically generating question-answer pairs to fit modern instruction-tuning formats. However, this approach is inherently bounded by the quality, limited semantic scope and nature of the underlying source data, failing to address the fundamental issues of scalability and the acquisition of novel domain knowledge.

\begin{figure}[!t]
    \centering
    \includegraphics[width=1\textwidth]{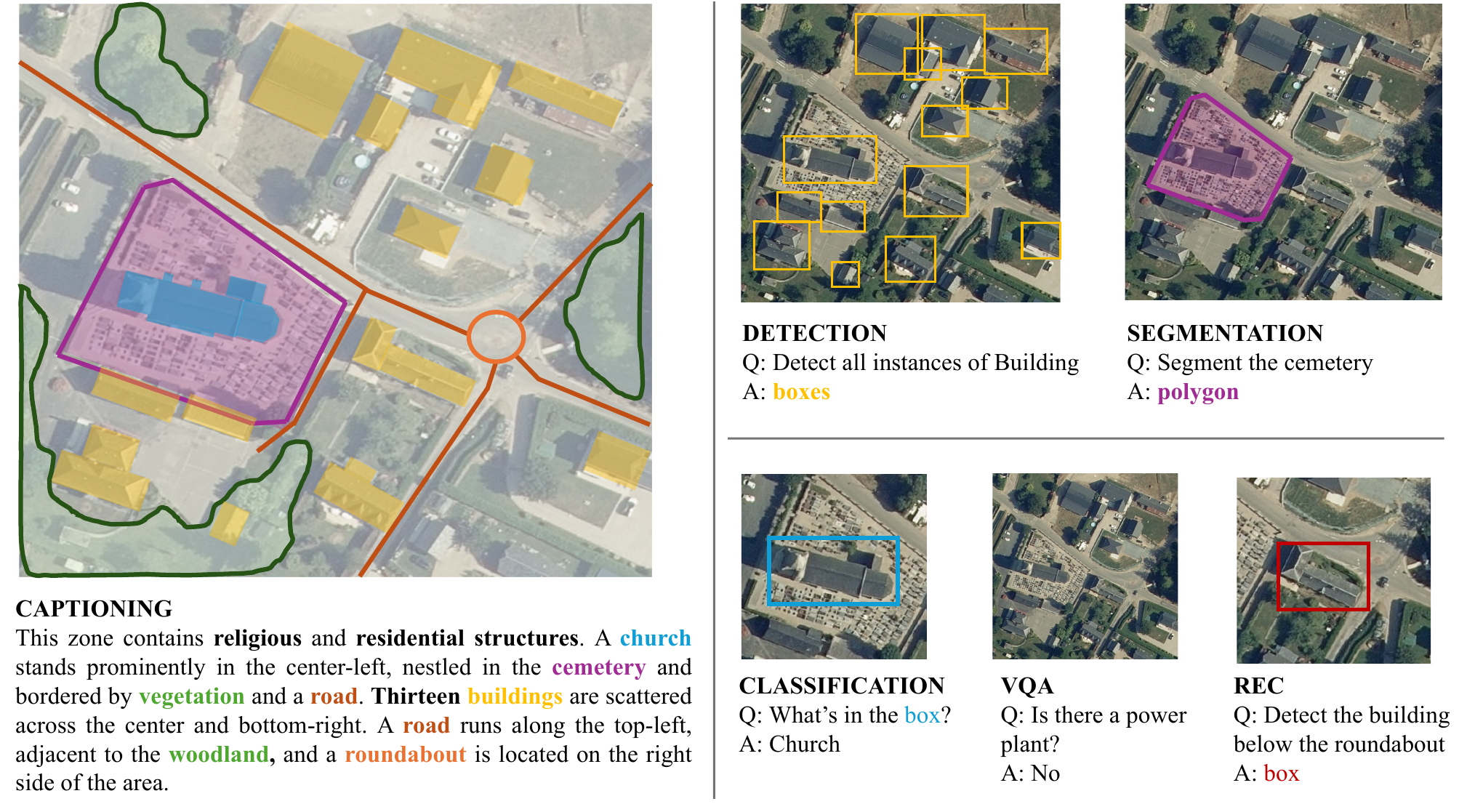}
    \caption{Overview of our dataset. We release both the raw vector data and a derived instruction-tuning dataset. The cadastral vector layers (e.g., buildings, vegetation, roads) serve as the ground truth and are automatically transformed into instructions for downstream tasks including scene captioning, object detection, semantic segmentation, localized classification, Visual Question Answering (VQA) and Referring Expression Comprehension (REC).
    \label{fig:figure}}
\end{figure}

In this work, we propose a novel, large-scale dataset comprising 3.8M annotated objects across 510k high-resolution images, specifically tailored for spatial understanding in EO. The primary innovation lies in our data curation methodology: rather than relying on legacy sets or noisy crowd-sourcing, we leverage cadastral vector data as our source of ground truth. This approach ensures the acquisition of high-fidelity geometric data that is manually curated and legally verified by national agencies, effectively solving the scalability-quality trade-off. Moreover, this rigorous sourcing allows us to introduce novel imagery while offering a semantic depth previously unavailable in the field. Whereas existing datasets are typically restricted to a few dozen generic classes (e.g., \cite{vrsbench, geochat, skysensegpt}), we provide 135 highly specific semantic labels. These include fine-grained categories such as power plants, heritage sites, specific crop types and diverse civil and industrial infrastructures. By exposing models to this level of administrative detail, we provide a challenging new benchmark for fine-grained visual recognition (Figure \ref{fig:figure}). Furthermore, we posit that this rigorous geometric foundation can support a broad spectrum of research applications beyond the tasks proposed in this paper (e.g., \cite{geolink, Bai2023GeographicMW, Audebert2017JointLF})

To empirically demonstrate the utility of this vector data, we establish a comprehensive instruction fine-tuning benchmark comprising seven spatial grounding tasks as a primary use case. Our experiments highlight a critical gap in the field: existing RS-specialized models (e.g., GeoChat \cite{geochat}, SkySenseGPT \cite{skysensegpt}) exhibit severe degradation in zero-shot generalization, sometimes underperforming relative to their base architectures. In contrast, we show that high-fidelity data enables standard architectures to master these tasks without complex architectural modifications. Our fine-tuned baseline significantly improves upon the zero-shot performance of specialized spatial models (e.g., Ferret \cite{ferret}) and commercial baselines. This performance gap serves as quantitative proof of the unique domain knowledge contained in our cadastral annotations—knowledge that is currently absent from web-scale pre-training.

The main contributions of this work are: 
\begin{enumerate} 
\itemsep0em 
\item \textbf{The Data:} A novel large-scale EO dataset distinguished by its verified cadastral origin, offering unprecedented semantic richness (135 classes) and density (3.8M objects).
\item \textbf{The Benchmark:} A comprehensive instruction fine-tuning suite comprising 7 spatial understanding tasks, derived from the vector data, to demonstrate its applicability to modern VLM training.
\item \textbf{The Baseline:} A fine-tuned model serving as a robust reference for future research, demonstrating that standard VLMs can master complex grounding tasks when trained on high-density verified vector data.
\end{enumerate}

\section{Related Work}
\label{sec:related}
\noindent
\textbf{General-Purpose Spatial VLMs}
Recent advances in spatial understanding rely on modular architectures like LLaVA \cite{llava} and PaliGemma \cite{paligemma}. To enable grounding, pioneering works such as Shikra \cite{shikra}, Kosmos-2 \cite{kosmos} and MiniGPT-v2 \cite{minigptv2} treat spatial coordinates as textual tokens, effectively bootstrapping spatial skills from pre-trained LLMs. Ferret \cite{ferret} further refined this by introducing a hybrid sampler for fine-grained feature extraction. These findings demonstrate the feasibility of embedding visual grounding directly into a single unified model, leveraging LLM's established reasoning and knowledge retrieval capabilities.

\noindent
\textbf{Remote Sensing MLLMs}
The Remote Sensing community adapted these paradigms via "continue pre-training" strategies. Early adopters GeoChat \cite{geochat}, SkySenseGPT \cite{skysensegpt}, EarthGPT \cite{earthgpt} and RSGPT \cite{HU2025272}, followed by recent works like GeoGround \cite{geoground} and GeoPixel \cite{geopixel}, fine-tune generic architectures on domain corpora. While validating RS VQA feasibility, these models remain bottlenecked by the quality and scale of training data.

\noindent
\textbf{Datasets} 
Data remains the primary constraint. Most benchmarks (e.g., VRSBench \cite{vrsbench}, GeoChat \cite{geochat}, GeoPixel \cite{geopixel}, RSVG \cite{10056343}) rely on legacy closed-set datasets like DOTA \cite{dota} and DIOR \cite{dior}, which are limited to generic categories. While large-scale initiatives like SkyScript \cite{skyscript} and SatLas \cite{satlas} rely on OpenStreetMap (OSM) for global coverage, they often trade off resolution and precision. In contrast, our work leverages high-resolution (20cm) cadastral data to provide the dense, verified vector annotations required for deep spatial understanding.

\section{Data}
\label{sec:data}
To ensure precise ground truth, we leverage cadastral records maintained by state administrations, which offer superior reliability compared to crowd-sourced alternatives. As a proof-of-concept, we utilize France's open-data infrastructure provided by the national mapping agency (IGN), combining high-resolution (20 cm) aerial imagery with legally verified vector data. Our pipeline is provider-agnostic, allowing seamless adaptation to other national standards (e.g., USGS, Ordnance Survey or GSI).

This design choice prioritizes annotation fidelity over immediate global diversity. While platforms like OSM provide expansive coverage, they often suffer from inconsistent taxonomy and geometric inconsistency. In contrast, we argue that for fine-grained spatial instruction tuning, the professional-grade cadastral data is the decisive factor for model performance.

Our dataset spans $85,864\text{ km}^2$ across 20 administrative departments, prioritizing urban density while ensuring diverse environments (alpine, maritime, rural). The optical aerial orthophotos, featuring a $20\text{cm}$ spatial resolution, are aligned with a map database containing vector models of a wide spectrum of stationary objects. These include buildings, transport infrastructure, land use zones and water bodies. Entities are modeled with $1\text{-meter}$ precision as (multi)polygons, lines or points. Crucially, annotations go beyond generic categories to include granular usage subtypes (e.g., distinguishing Catholic from Orthodox churches), enabling highly specific semantic recognition.

\subsection{Pipeline}
From the selected departments, we extracted over 4 million patches. Each patch has a size of $672 \times 672$ pixels, corresponding to a spatial extent of approximately $134 \times 134$ meters. These patches underwent a rigorous processing pipeline designed to eliminate redundancy, filter categories, reduce noise and generate instruction sets.

\subsubsection{Preprocessing} 
\begin{figure}[!p]
    \centering
    \includegraphics[width=1\textwidth]{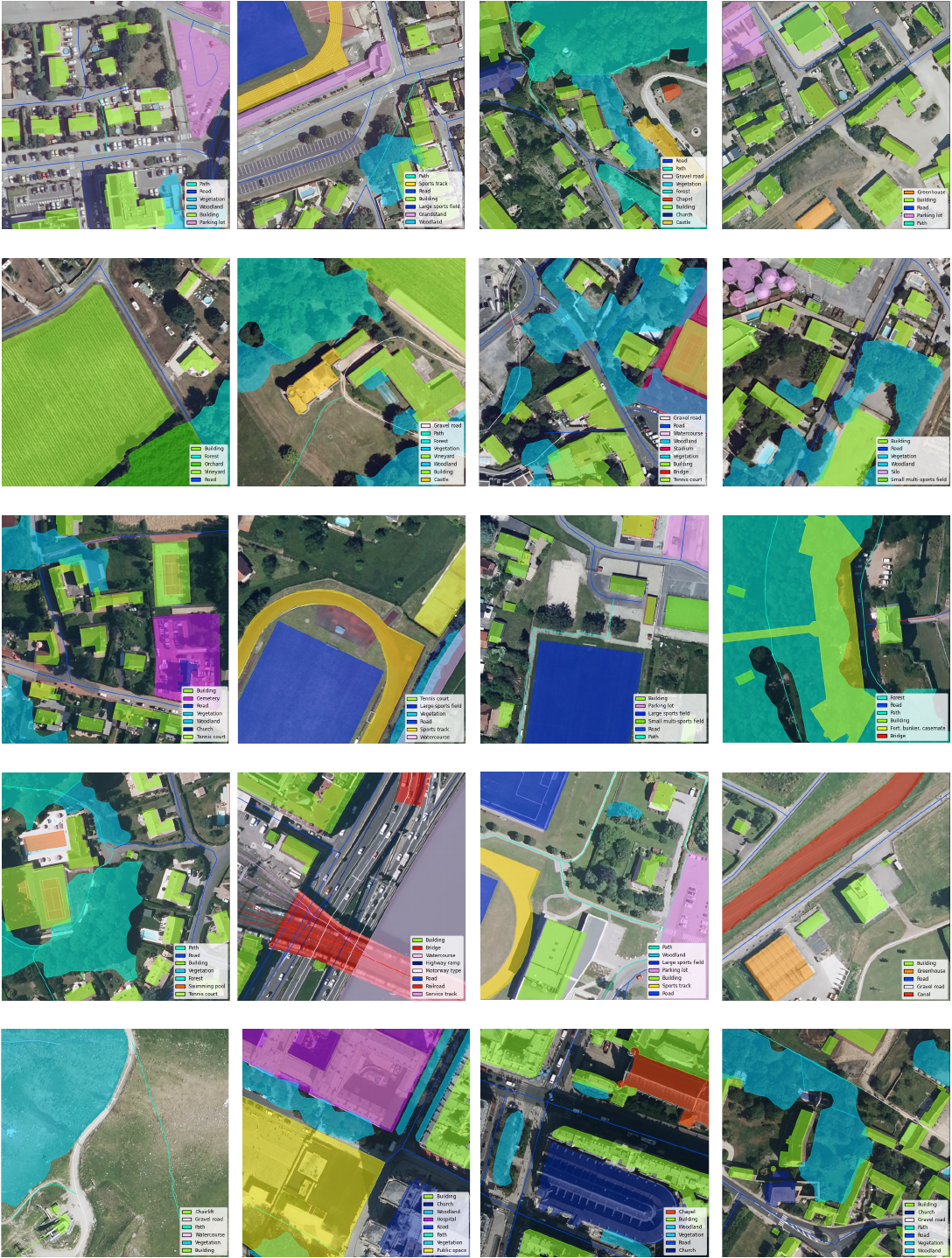}
    \caption{Qualitative samples from the proposed dataset, illustrating the semantic granularity and geometric precision of the cadastral annotations across diverse environments.}
    \label{fig:examples}
\end{figure}

Initially, we filter entities that are not visually discernible in orthophotos, such as tunnels, subterranean conduits or underground parking structures. This ensures strict alignment between visual features and vector data, preventing downstream hallucinations. The 735 unique semantic terms in the database were automatically translated into English and manually verified to serve as ground-truth labels. See Figure \ref{fig:examples} for some qualitative examples.

Given the sparsity of rural territories, a significant portion of raw scenes contain minimal information or repetitive patterns. The scene composition follows a long-tail distribution, dominated by single-element images (e.g., uniform forests) or highly frequent combinations. To mitigate this semantic redundancy, we applied a resampling strategy: we group images based on their unique set of present entities (e.g., \{`Forest', `Path'\}) and capped each unique combination at a maximum of 64 samples. This strategy reduced the dataset size by 85\% while effectively balancing the distribution of scene compositions.

The raw cadastral data often exhibits administrative granularity misaligned with visual semantics. For instance, a single functional structure (e.g., a residence and its detached garage) or a dense urban block may be legally partitioned into multiple units despite appearing as a continuous entity. To solve this issue, we merge contiguous polygons of the same class to ensure the vector ground truth reflects the visual footprint rather than legal partitions, which are often indiscernible.

\begin{figure}[!h]
    \centering
    \begin{subfigure}[b]{0.48\textwidth}
        \centering
        \includegraphics[width=\textwidth]{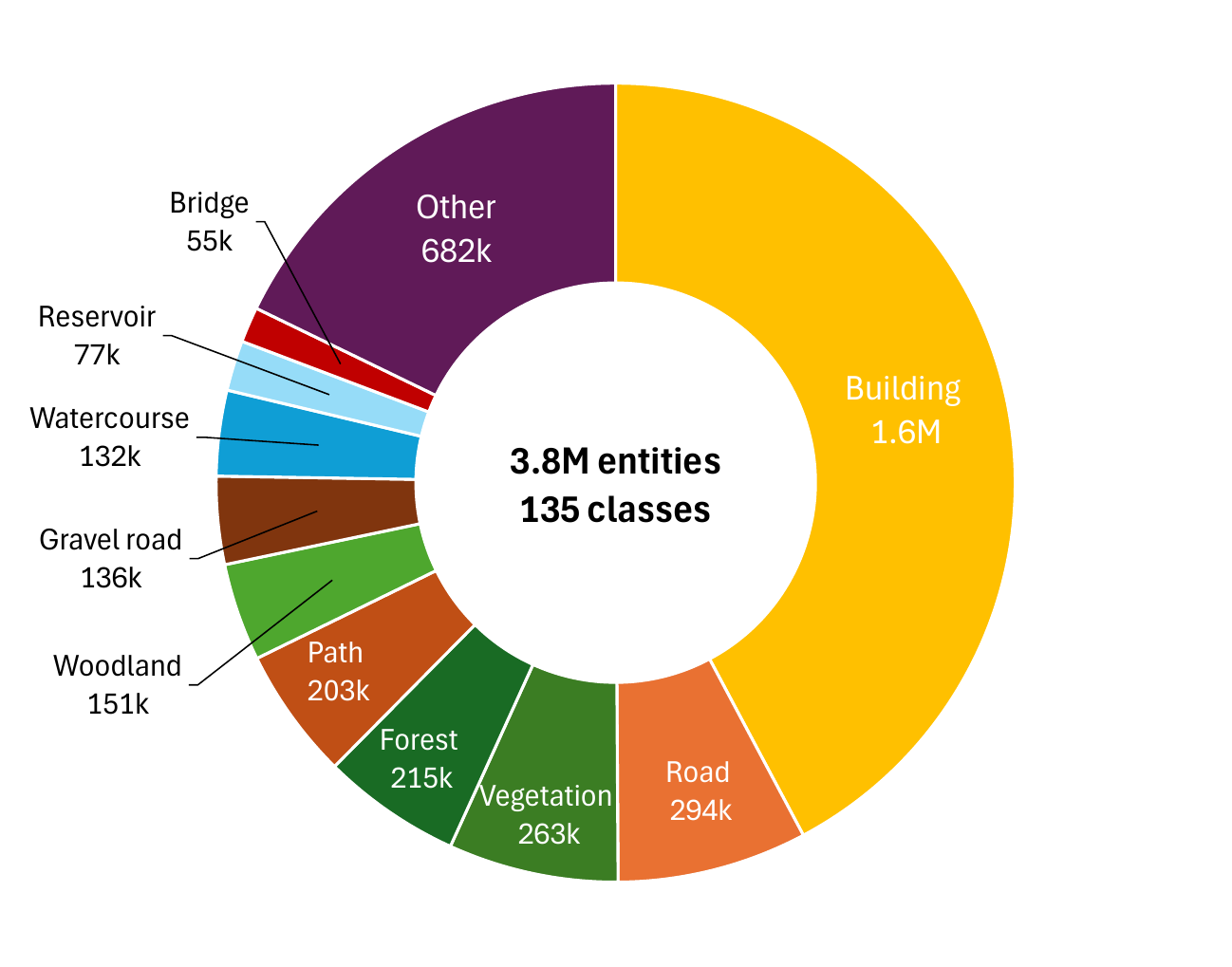}
        \caption{Occurrences of the top-10 classes.}
        \label{fig:classes}
    \end{subfigure}
    \hfill 
    \begin{subfigure}[b]{0.43\textwidth}
        \centering
        \includegraphics[width=\textwidth]{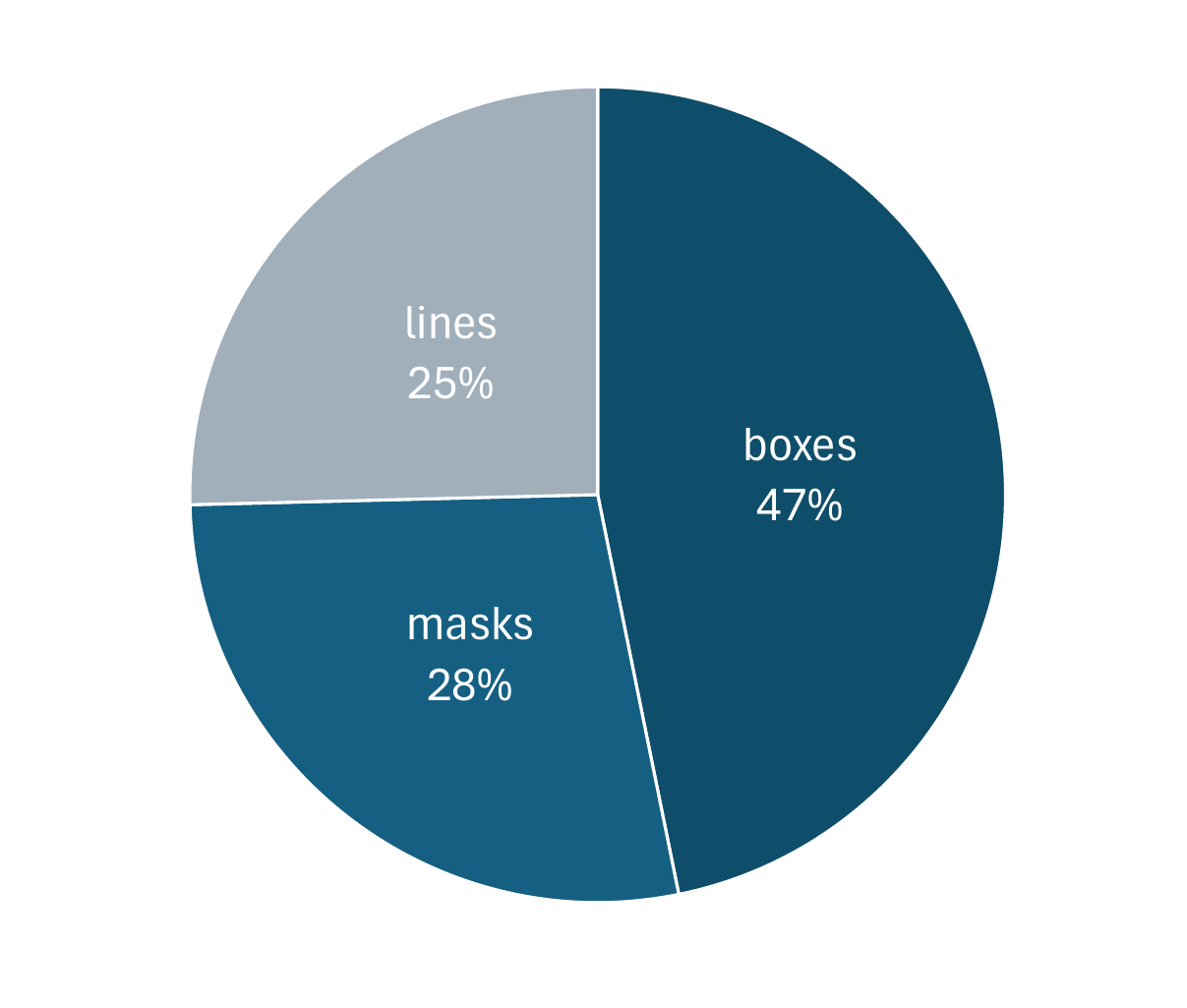}
        \caption{Geometric annotations in the dataset.}
        \label{fig:annotations}
    \end{subfigure}
    \caption{The data covers 135 unique semantic categories: a) beyond the top-10, common classes include Canals (44k), Railroads (40k), Parking lots (16k), Tennis courts (15k), Cemeteries (13k) and Stadiums (10k). In terms of geometry, b) the dataset comprises 3.8M entities including linear features (e.g., roads, railroads, canals), bounding boxes (derived from original GIS polygons) and segmentation masks.}
    \label{fig:pies}
\end{figure}

Furthermore, padding is applied to exclude small or partial objects cropped at image borders which lack sufficient context for recognition. Finally, geographical coordinates are normalized into relative pixel coordinates in the range $[0, 1]$. After the preprocessing phase, the dataset comprises 510,483 patches containing 3,829,755 objects across 135 unique categories (Figure \ref{fig:pies}).

\subsubsection{Annotations}
While the original vector data provides polygons, lines and points, we augment these with Horizontal Bounding Boxes (HBB) and Oriented Bounding Boxes (OBB) to support diverse downstream tasks. Segmentation masks are represented as coordinate sequences $[x_1, y_1, \dots, x_N, y_N]$. HBBs are defined as $[x_{min}, y_{min}, x_{max}, y_{max}]$, while OBBs are represented as $[x_{center}, y_{center}, h, w, \theta]$. Elements with expansive or irregular geometries (e.g., airports, rivers, extensive vegetation) are assigned to a "segmentation-only" category to avoid ambiguous or too wide bounding boxes.

Our dataset maintains a high annotation density (median: 8 objects/image), significantly outperforming standard benchmarks like VRSBench (median: 2). While GeoPixel exhibits a comparable median (9), its density is largely driven by transient objects (e.g., vehicles), whereas our density is derived exclusively from stationary cadastral elements. Furthermore, unlike prior baselines that approximate polygons from raster masks, we utilize native vector data. This ensures precise, human-verified geometric primitives that are topologically accurate and inherently more efficient for sequence-based modeling.

In addition to geometric annotations, we generate descriptive text to synthesize scene information. To ensure scalability, this process is fully automated using state-of-the-art VLMs. However, unlike prior benchmarks (e.g., \cite{vrsbench, skysensegpt, geopixel}) that task VLMs with interpreting visual attributes or background context -- effectively distilling possible biases into the dataset -- we strictly limit the VLM to elaborate on pre-defined raw captions. We ground the generation of raw captions on rigid geometric rules and templates, constraining the model to merely rephrase existing facts rather than perceive new ones. This prevents the hallucination of non-existent objects or attributes while allowing for the inclusion of novel semantic categories. 

\subsubsection{Visual-Semantic Discrepancy} 
While cadastral data provides high-fidelity ground truth, the alignment between abstract vector records and optical imagery might contain some discrepancies. We estimate a residual divergence rate comprising: $\sim$3\% errors (i.e., annotations without visual evidence, due for example to temporal misalignment); $\sim$5\% incomplete records (i.e., visible elements lacking geometry in the cadastre); and $\sim$7\% occluded instances (i.e., objects partially or fully covered by vegetation/shadows). Crucially, we argue that the latter category represents a realistic operational challenge rather than noise. In real-world scenarios, detecting an object requires inferring its existence from context (e.g., a driveway disappearing into a forest patch) even when not visible.

To systematically characterize these visual-semantic discrepancies, we implement an automated profiling pipeline using \textit{Gemini-2.5 Flash}. Each annotation is visually highlighted (via Set-of-Marks prompting \cite{yang2023setofmark}) and submitted for verification, where the model is tasked solely to confirm the visual presence of the element.
Crucially, instances deemed "invisible" are flagged rather than discarded. This approach preserves the integrity of the original cadastral record while providing downstream users with the flexibility to select between "clean" (visually confirmed) and "hard" (occluded/inferred) training splits.

\subsubsection{Postprocessing}
The high granularity of cadastral metadata required semantic simplification for an effective learning. Subtypes lacking distinctive visual features (e.g., \textit{primary} vs. \textit{high school}, \textit{shrubby} vs. \textit{herbaceous heath}) were aggregated into their super-categories (e.g., \textit{School}, \textit{Vegetation}). Conversely, visually distinct functional categories (e.g., \textit{hospitals}, \textit{greenhouses}) were retained.

The resulting label space is hierarchical and not mutually exclusive. A query for a generic parent class (e.g., \textit{Building}) must accept all valid subtypes (e.g., \textit{church}, \textit{train station}) as positive instances. To address this, we release a comprehensive taxonomy to guide training and evaluation. 

\subsection{Dataset}
\label{sec:dataset}
We provide two distinct data releases: a large-scale pre-training dataset (3.8M objects, 510k images) and a curated VQA dataset tailored for instruction fine-tuning (880k objects, 60k images, 1.8M questions). While this work focuses on the latter to benchmark models and fine-tune our baseline, the full-scale release is intended to support broader research beyond the MLLM paradigm. A comparison with existing benchmarks is provided in Table \ref{tab:datasets}.

\begin{table}[h]
\centering
\begin{tabular}{lcccccc}
\toprule
\textbf{Dataset} & \textbf{Tasks} & \textbf{Annotations} & \textbf{Classes} & \textbf{Objs.} & \textbf{Imgs.} & \textbf{Instructions} \\ 
\midrule
VRSBench \cite{vrsbench}   & 3/7 & HBB            & 26  & 52k  & 29k & 205k \\
GeoPixel \cite{geopixel}    & 2/7 & Poly           & 15  & 662k   & 19k & 53k \\
GeoChat \cite{geochat}    & 4/7 & OBB            & 45  & --   & 140k & 318k \\
SkySenseGPT \cite{skysensegpt} & 5/7 & OBB            & 48  & 210k & 82k & \textbf{1.8M} \\ 
\midrule
\textbf{GroundSet (full)} & 4/7 & \textbf{HBB, OBB, Poly} & \textbf{135} & \textbf{3.8M} & \textbf{510k} & -- \\
\textbf{GroundSet (sft)}   & \textbf{7/7} & HBB, Poly      & 126 & 880k & 60k & \textbf{1.8M} \\ 
\bottomrule
\end{tabular}
\caption{Comparison with existing benchmarks across seven core tasks: captioning ~\cite{vrsbench, geopixel, geochat, skysensegpt}, classification ~\cite{geochat, skysensegpt}, object detection ~\cite{skysensegpt} and multi-class detection, segmentation ~\cite{geopixel}, referring expression comprehension ~\cite{vrsbench, geochat, skysensegpt} and VQA ~\cite{vrsbench, geochat, skysensegpt}. Our contribution includes a large full release with detailed geometry and a smaller supervised fine-tuning subset focused on instructions.}
\label{tab:datasets}
\end{table}

\subsubsection{Tasks} 
In our benchmark, we define seven tasks to assess spatial understanding capabilities (Figure \ref{fig:figure}).
\begin{itemize} 
\item \textbf{Captioning:} generate a coherent paragraph describing the scene. 
\item \textbf{Localized Classification:} classify a given region (could be a box or complex polygon).
\item \textbf{Object Detection:} localize specific classes using Horizontal Bounding Boxes (HBB). 
\item \textbf{Multi-Class Detection:} localize instances from a list of multiple target classes. 
\item \textbf{Referring Expression Comprehension (REC):} localize an object based on a textual description. 
\item \textbf{Segmentation:} localize target classes using polygonal masks. 
\item \textbf{Visual Question Answering (VQA):} binary verification of object presence. 
\end{itemize}

\subsubsection{Instructions} 
We partition the finetuning dataset into 14,000 images for testing and 45,988 images for training. While we provide a comprehensive instruction-tuning dataset generated via templates to train our baseline, we emphasize that the release of the underlying raw vector data enables significantly broader methodological exploration. We encourage the research community to leverage these native geometric primitives to design novel training paradigms that extend beyond the specific instruction formats presented here. Conversely, to ensure consistent and comparable benchmarking, the test set questions are fixed.

Test Set: Evaluation instructions are generated by integrating annotations into diverse templates (10 linguistic variations per question) to ensure generalization. To evaluate robustness against hallucination, we incorporate negative samples (21\% probability) into detection and segmentation tasks, prompting the model to locate absent objects. The test set comprises 72,597 instructions distributed across the tasks.

Training Set: For our baseline, we extend the standard task instructions with auxiliary objectives to augment training data. These include exhaustive segmentation queries for every annotated object and text-only geometric reasoning tasks. The latter utilize synthetic random polygons, asking the model to: 1) simplify shapes by removing vertices while preserving topology, and 2) predict bounding boxes for the polygon inputs. These auxiliary tasks are designed to acclimatize the model to the coordinate system (discretized in the interval $[0-1000]$). The training set contains in total 1,845,076 instructions.

\section{Evaluation}
\label{sec:eval}
Given the newly acquired imagery and novel semantic categories introduced by our dataset, we benchmark state-of-the-art models in a strictly zero-shot setting to assess generalization. We evaluate a diverse suite of models: 
\begin{enumerate} 
\itemsep0em 
\item Open-Source Baselines: \textit{LLaVA-1.5} \cite{llava} and \textit{LLaVA-1.6} \cite{liu2024llavanext}. 
\item Remote Sensing Specialists: \textit{GeoChat} \cite{geochat}, \textit{SkySenseGPT} \cite{skysensegpt} and a \textit{VRSBench} baseline\footnote{We fine-tune LLaVA-1.5 on VRSBench \cite{vrsbench} with the same hyper-parameters reported in their paper.}. 
\item Spatial Grounding Specialists: \textit{Ferret} \cite{ferret}, \textit{MiniGPT-v2} \cite{minigptv2} and \textit{PaliGemma-2} \cite{paligemma}. 
\item Commercial Baseline: \textit{Gemini-2.5 Flash}, representing native multimodal capabilities at scale. 
\end{enumerate}

To ensure a fair comparison, we select open-source models with comparable parameter counts (approx. $7\text{B}$), with the exception of PaliGemma ($10\text{B}$). While architectural nuances exist, 
they all share a modular design philosophy that contrasts with the native multimodal architecture of Gemini. Crucially, all Remote Sensing specialists -- including our own fine-tuned baseline -- are based on the LLaVA architecture. By fixing the model architecture, we effectively isolate the training data as the primary variable, allowing us to rigorously assess the impact of our proposed dataset against existing alternatives.

\subsection{Metrics} 
\label{sec:metrics} 
We adapt our evaluation protocols to accommodate the diverse requirements of heterogeneous models in a zero-shot setting. On the output side, we strictly do not penalize formatting deviations; predictions are geometrically recovered (e.g., converting OBB to HBB or rescaling coordinate norms) prior to scoring. On the input side, standard conversational instructions often yields poor zero-shot performance. Consequently, we tailor the input prompts to match the specific template expected by each architecture\footnote{For instance, PaliGemma expects specific task prefixes (e.g., \texttt{detect} followed by the object name) rather than conversational instructions.}. Furthermore, some models are exclusively trained on OBB thus interpreting the HBB as oriented boxes with $90^{\circ}$ angle. Thus, questions are systematically adapted per-model to ensure a fair assessment of their true spatial grounding capabilities.

\textbf{Captioning.} We report standard Natural Language Generation metrics: BLEU@4, METEOR and CIDEr. To prevent score distortion, we filter bounding box tokens and unwanted artifacts generated by specialized models prior to evaluation.

\textbf{Classification.} Alongside exact string matching, we employ a Semantic Similarity metric for this open-vocabulary task. Predictions are normalized and, if embedded in verbose sentences\footnote{(e.g., "The label that applies is `bridge'" or "This image likely contains a parking lot")}, extracted using an LLM (\textit{Gemma-3}) to isolate the named entity. A prediction is deemed correct if its embedding (computed via \texttt{all-mpnet-base-v2}) exceeds a similarity threshold $k$ with the ground truth (Acc@k), robustly handling synonyms (e.g., \textit{"tennis court"} vs. \textit{"tennis field"}).

\textbf{Grounding.} For Object Detection, Multi-Class Detection, Referring Expression Comprehension and Segmentation, we compute mIoU regardless of the output modality (HBB, OBB or mask). Crucially, evaluation is taxonomy-aware: predicting a valid child class (e.g., \textit{Church}) when queried for a parent class (e.g., \textit{Building}) is considered correct. We select the candidate with maximum IoU per target and report F1, Precision and Recall based on:
\begin{itemize}
\itemsep0em
\item \textbf{TP:} IoU $\ge 0.5$ with an unclaimed ground truth.
\item \textbf{FP:} IoU $< 0.5$ or duplicate prediction for the same ground truth.
\item \textbf{FN:} Unmatched ground truth object.
\item \textbf{TN:} Correct rejection (no object predicted when none exists).
\end{itemize}

\textbf{Visual Question Answering (VQA).} Treated as binary classification (Presence/Absence). We report Precision, Recall, F1 and Accuracy after standard text normalization.

\subsection{Finetuning} To establish a robust baseline for this benchmark, we fine-tune LLaVA-1.6 7B  on the proposed instruction set. We select this architecture as the canonical representative of modular MLLMs, a design paradigm adopted by prominent domain-specific models such as GeoChat and SkySenseGPT\footnote{We do not fine-tune RS-specialists because their shared LLaVA architecture would conflate legacy pre-training with GroundSet, preventing a strict data ablation.}. Crucially, LLaVA-1.6 is distinguished by its dynamic resolution capability (AnyRes), which is particularly suitable for processing our high-resolution aerial imagery. 

Concerning the implementation, we adhere to a parameter-efficient fine-tuning strategy. The model is trained for a single epoch using LoRA (Low-Rank Adaptation) \cite{hu2022lora} without the introduction of auxiliary spatial modules or architectural modifications. Training was conducted on $8 \times \text{A100}$ GPUs over approximately 72 hours. By deploying the vanilla architecture, we aim to isolate the contribution of the data from architectural priors, thereby providing a clean and reproducible reference point for future research.

\subsection{Results}
We benchmark the models on the test set described in Section \ref{sec:dataset}. To ensure computational efficiency, all models were evaluated using 4-bit quantization, following preliminary experiments confirming negligible performance impact. We summarize the results in Table \ref{tab:results}, where the \textit{RS-ft} column distinguishes between models fine-tuned on Remote Sensing corpora and general-purpose baselines. Each RS-specialist is trained on its own dataset and evaluated zero-shot on our test set. This comparison isolates the impact of training data.

\begin{table}[h]
\centering
\small
\setlength{\tabcolsep}{2pt}
\begin{tabular*}{\textwidth}{@{\extracolsep{\fill}} l c ccccccc @{}}
\toprule
 & \textbf{RS} & Cap. & VQA & Class. & Det. & Mult. & REC & Seg. \\
 & \textbf{ft}  & (CIDER) & (F1) & (acc@0.8) & (F1@0.5) & (F1@0.5) & (F1@0.5) & (F1@0.5) \\
\midrule
\midrule
Gemini-2.5 & \textcolor{red}{$\times$} & 16.68 & 83.56 & \uline{49.84} & 3.79 & 9.98 & 4.70 & 2.79 \\
\midrule
LLaVA-1.5 & \textcolor{red}{$\times$} & 18.74 & 87.01 & 28.62 & 1.48 & 2.96 & 2.71 & 0.20 \\
LLaVA-1.6 & \textcolor{red}{$\times$} & 17.93 & 86.36 & 29.20 & 0.92 & 3.96 & 4.10 & 0.07 \\
\midrule
MiniGPT-v2 & \textcolor{red}{$\times$} & 0.24 & 76.41 & 6.97 & 11.69 & 18.48 & \uline{17.47} & - \\
PaliGemma-2 & \textcolor{red}{$\times$} & 18.04 & 85.34 & - & 17.85 & 20.42 & - & 15.12 \\
Ferret & \textcolor{red}{$\times$} & 18.73 & 68.20 & 15.34 & \uline{18.08} & \uline{32.83} & 13.87 & \uline{19.77} \\
\midrule
GeoChat & \textcolor{green}{\checkmark} & 3.99 & 88.96 & 7.59 & 5.52 & 5.84 & 6.13 & - \\
SkySenseGPT & \textcolor{green}{\checkmark} & \uline{24.03} & \uline{89.69} & 8.15 & 3.13 & 0.14 & 1.63 & - \\
VRSBench & \textcolor{green}{\checkmark} & 2.65 & 87.22 & - & 2.33 & 7.26 & 3.25 & - \\
\midrule
GroundSet & \textcolor{green}{\checkmark} & \textbf{104.48} & \textbf{97.09} & \textbf{94.18} & \textbf{49.47} & \textbf{72.14} & \textbf{39.45} & \textbf{44.65} \\
\bottomrule
\end{tabular*}
\caption{Zero-shot evaluation of state-of-the-art models compared to Remote Sensing fine-tuned baselines (RS-ft) and our new baseline which is trained on the proposed dataset. We observe that: (i) General spatial models (Ferret, PaliGemma) surpass RS-specialists in their own domain; (ii) Previous RS models might underperform their foundational baselines and do not generalize well on our benchmark; and (iii) Commercial-grade solutions (Gemini) show strong general competence but ultimately lack the specific semantic knowledge to close the gap with our fine-tuned model.}
\label{tab:results}
\end{table}

\textbf{Captioning.} Given the strict structural constraints of our dataset's ground truth, only our fine-tuned baseline achieves fully satisfactory alignment. SkySenseGPT marginally outperforms other models, which cluster around similar baseline values.

\textbf{Classification.} In the classification task, a distinct trend emerges: general-purpose models significantly outperform specialized RS models. This suggests the strategy employed by models like SkySenseGPT and GeoChat induces catastrophic forgetting, eroding zero-shot generalization. Conversely, larger commercial models benefit from extensive pre-training on diverse data. However, a critical gap remains: our fine-tuned model surpasses the strongest commercial baseline (Gemini) by a mean margin of 44 points. This substantial gap underscores the unique value of our dataset, which introduces semantic categories and visual distributions absent from web-scale training corpora.

\textbf{Visual Question Answering (VQA).} The VQA performance landscape is  more nuanced. Unlike classification, this task requires less precise localization, reducing the penalty for models with weak spatial priors. Consequently, RS-specialized models demonstrate a slight advantage over generic baselines due to domain adaptation.

\textbf{Grounding Tasks.} For tasks demanding high-fidelity spatial grounding -- such as Object Detection, Multi-Class Detection and Segmentation --architectural priors provide some benefits. Specialized architectures like Ferret (with its hybrid region sampler) and PaliGemma significantly outperform both RS-specialized models and commercial APIs in zero-shot settings. Notably, Ferret achieves the highest zero-shot detection scores, suggesting that specific solutions for general purpose localization are more impactful than domain-specific pre-training alone.

\textbf{The Data Factor.} We utilize a standard LLaVA-1.6 architecture to strictly isolate the impact of training data from architectural priors. Since leading RS-specialists (e.g., GeoChat, SkySenseGPT, VRSBench) are built upon the same LLaVA meta-architecture, they effectively serve as control baselines for legacy training data. The substantial performance disparity observed in Table \ref{tab:results} is therefore not a result of superior model design, but is directly attributable to the semantic richness and geometric precision of our cadastral annotations. This result exposes a critical "knowledge gap" in prior datasets and demonstrates that standard VLMs can master fine-grained spatial understanding when grounded in high-fidelity data, without requiring complex architectural modifications.

\textbf{Training mix.}
We validate our training strategy with ablation studies on a 100k-instruction subset, as detailed in Table \ref{tab:ablation}. The results demonstrate that augmenting the training mixture with auxiliary geometric tasks -- specifically text-only polygon reasoning and exhaustive segmentation queries described in Section \ref{sec:dataset} -- improves spatial grounding capabilities. While these augmentations induce a negligible drop in classification accuracy, they drive meaningful gains in detection and segmentation scores. In contrast, unfreezing the vision encoder yields inconsistent results, notably degrading segmentation performance. Finally, we observe that filtering occluded `invisible' instances improves the segmentation scores, albeit at a slight cost in other tasks. Based on these findings, we maintain a frozen vision encoder and incorporate both geometric augmentation strategies in our final training recipe.

\textbf{Cross-Dataset Generalization.}
To assess robustness beyond our specific data distribution, we evaluate our GroundSet-trained baseline on the VRSBench benchmark in a zero-shot setting. Since the original release is limited to captioning, REC and VQA, we expanded the benchmark by generating new questions for Object Detection directly from the underlying annotations. This evaluation represents a severe out-of-distribution test: VRSBench is heavily tilted toward mobile objects (e.g., vehicles, ships, airplanes) categories strictly absent from our stationary cadastral training set. 

Consequently, we observe an uneven transfer across different task types. As shown in Table 4, our model trails RS-specialists in tasks such as Captioning and VQA. We attribute this to the fact that baseline models are trained with differing task definitions; consequently, performance on these tasks is highly sensitive to domain-specific vocabulary and rigid prompt templates. In zero-shot settings, even minor deviations from the expected prompt structure can degrade performance on these language-dependent tasks.

Despite this domain shift, a remarkable trend emerges in the core spatial grounding tasks. Our model outperforms leading RS-specialists in both Detection and Referring Expression Comprehension. This highlights a striking performance asymmetry: while existing RS-specialists collapse when evaluated zero-shot on our dataset (Table \ref{tab:results}), our model preserves robust grounding capabilities when transferred to VRSBench. This achievement is particularly notable considering that prior RS-specialists were pre-trained on DOTA and DIOR imagery -- the source domains of VRSBench -- giving them an in-domain visual advantage, whereas our baseline operates in a strictly out-of-distribution setting. Ultimately, these results clarify that while out-of-the-box cross-domain robustness on language-heavy tasks requires downstream fine-tuning, GroundSet is highly effective at instilling foundational visual grounding. While semantic vocabulary is domain-dependent, the core geometric understanding acquired from our dense vector data successfully transfers to new domains.

\begin{table}[hbt!]
    \centering
    \begin{tabular}{lccc}
        \toprule
        & Class. & Det. & Seg. \\
        & (acc@0.8) & (F1@0.5) & (F1@0.5) \\
        \midrule
        Baseline & \textbf{90.47} & 12.77 & 12.62 \\
        Text Aug. & \underline{90.06} & \textbf{13.39} & 14.52 \\
        Seg. Aug. & 89.95 & \underline{13.28} & \underline{14.54} \\
        ViT Unfrozen & 89.93 & 13.12 & 11.52 \\
        Filtered & 89.99 & 11.95 & \textbf{15.31} \\
        \bottomrule
    \end{tabular}
    \caption{Ablation results (100k subset).Integrating text-only geometric reasoning (\textit{Text Aug.}) and dense segmentation queries (\textit{Seg. Aug.}) boosts spatial task performance over the baseline, while unfreezing the visual encoder proves detrimental. Filtering invisible objects (\textit{Filtered}) significantly improves segmentation at the cost of detection.}
    \label{tab:ablation}

    \vspace{0.5cm}

    \begin{tabular}{lccccc}
        \toprule
        & Cap. & VQA & Det. & REC \\
        & (CIDEr) & (F1) & (F1@0.5) & (F1@0.5) \\
        \midrule
        LLaVA-1.5 & 21.16 & 91.54 & 3.45 & 9.96 \\
        LLaVA-1.6 & 6.75 & 81.85 & 2.84 & 11.39 \\ \midrule
        GeoChat & 17.18 & 87.97 & 9.20 & 12.07 \\
        SkySenseGPT & 24.36 & 90.35 & 12.81 & 1.67 \\
        VRSBench & \textbf{45.55} & \textbf{92.36}  & 4.38 & 9.46 \\
        \midrule
        GroundSet & 12.97 & 68.03  & \textbf{15.47} & \textbf{21.49} \\
        \bottomrule
    \end{tabular}
    \caption{Cross-Dataset Evaluation (Zero-Shot on VRSBench). Our GroundSet-trained model outperforms RS-specialists in core grounding tasks (Det. and REC). Despite operating strictly out-of-distribution (missing vehicles/planes) against competitors with in-domain pre-training advantages on the source imagery, our baseline demonstrates superior cross-domain spatial understanding.}
    \label{tab:cross}
\end{table}

\section{Conclusion}
In this work, we addressed the critical scarcity of high-fidelity spatial annotations in Remote Sensing by introducing a large-scale dataset grounded in verifiable cadastral records. Comprising 3.8 million objects across 510k high-resolution images, this release provides a level of semantic granularity (135 classes) and geometric precision previously unavailable in the field. Unlike synthetic or crowd-sourced alternatives, our data offers legally verified ground truth, establishing a rigorous foundation for spatial AI research.

To validate this resource, we benchmarked state-of-the-art models on seven spatial tasks. Our experiments demonstrate that fine-tuning a standard, unmodified VLM on this data is sufficient to acquire robust spatial grounding capabilities that are currently beyond the reach of specialized architectures and massive commercial baselines. This result highlights a critical "knowledge gap" in current web-scale models and confirms that our dataset encapsulates new specific and granular domain knowledge.

\textbf{Future Work.} We release both the instruction dataset and the raw vector data to foster further methodological innovation. While our autoregressive baseline establishes a new state-of-the-art, the release of native vector primitives uniquely positions the community to explore entirely new modeling paradigms. We encourage future research to leverage these geometric structures to investigate non-autoregressive architectures -- such as GNNs, JEPA-inspired architectures or diffusion LLMs -- to go beyond the standard strategies of current VLMs.

\section*{Acknowledgements}
This work was supported by Google under a research collaboration agreement with Université Paris Cité. We also gratefully acknowledge the support of the Google Cloud Platform (GCP) and HPC resources from GENCI–IDRIS (Grant 2025-AD011016789).

\clearpage

\bibliographystyle{splncs04}
\bibliography{main}

\clearpage
\appendix

\section{Supplementary Material}
\section{Repository Structure}
The repository provides the data, model weights and the code to explore and evaluate the proposed \textit{GroundSet} dataset.

\begin{tcolorbox}[colback=white, colframe=gray!50, title=\textbf{Repository}]
\small
\dirtree{%
.1 GroundSet/.
.2 pretraining/ \dotfill \textit{Pretraining dataset}.
.3 images/ \dotfill \textit{Aerial images}.
.3 json/ \dotfill \textit{Raw annotations (JSON)}.
.2 finetuning/ \dotfill \textit{Supervised finetuning dataset}.
.3 images/.
.3 json/ \dotfill \textit{JSON files with train/test split}.
.2 instructions/ \dotfill \textit{Instructions for SFT (train/test)}.
.2 model/ \dotfill \textit{Pretrained weights}.
.2 src/ \dotfill \textit{Codebase}.
.3 eval/ \dotfill \textit{Evaluation Scripts}.
.4 eval\_caption.py.
.4 eval\_detection.py.
.4 [\dots].
.3 inference/ \dotfill \textit{VLM Inference Code}.
.4 llava.py.
.4 gemini.py.
.4 ferret.py.
.4 [\dots].
.3 resources/.
.4 tree.json \dotfill \textit{Dataset Taxonomy}.
.3 train/ \dotfill \textit{Finetuning Code}.
.4 finetune.py.
.4 inference.py.
.4 merge.py.
.4 zero2.json.
}
\end{tcolorbox}

\warning \textbf{Note:} To prevent data leakage, the pretraining dataset does not contain the samples used in the finetuning subset. 

\newpage
\section{Dataset Details}

\subsection{Data Format}
\textbf{1. Raw Data.} Each JSON file contains image metadata and grounding annotations.
\begin{lstlisting}[language=json, mathescape=true]
{
  "id": "13-2020-0805-6275-LA93-0M20-E080_419",
  "caption": "This is a rural area characterized by...",
  "image": "BDOrtho_13-2020-0805-6275.png",
  "source": "BDTOPO",
  "resolution": [672, 672],
  "objects": {
    "Building1": {
      "class": "Building",
      "mask": [[$x_1, y_1, x_2, y_2, \dots, x_n, y_n$]],
      "hbb": [[$x_{min}, y_{min}, x_{max}, y_{max}$]],
      "geometry": null
    }
  }
}
\end{lstlisting}

\noindent \textbf{2. Instructions.} Question-answer pairs are stored in JSONL format:
\begin{lstlisting}[language=json, mathescape=true]
{"id": "207_2", "image": "BDOrtho_207.png", "question": "...", "answer": "...", "class": "..."}
\end{lstlisting}

\subsection{Statistics and Coverage}
\begin{figure}[hbt!]
    \centering
    \begin{minipage}[c]{0.42\textwidth} 
        \centering
        \includegraphics[width=\textwidth]{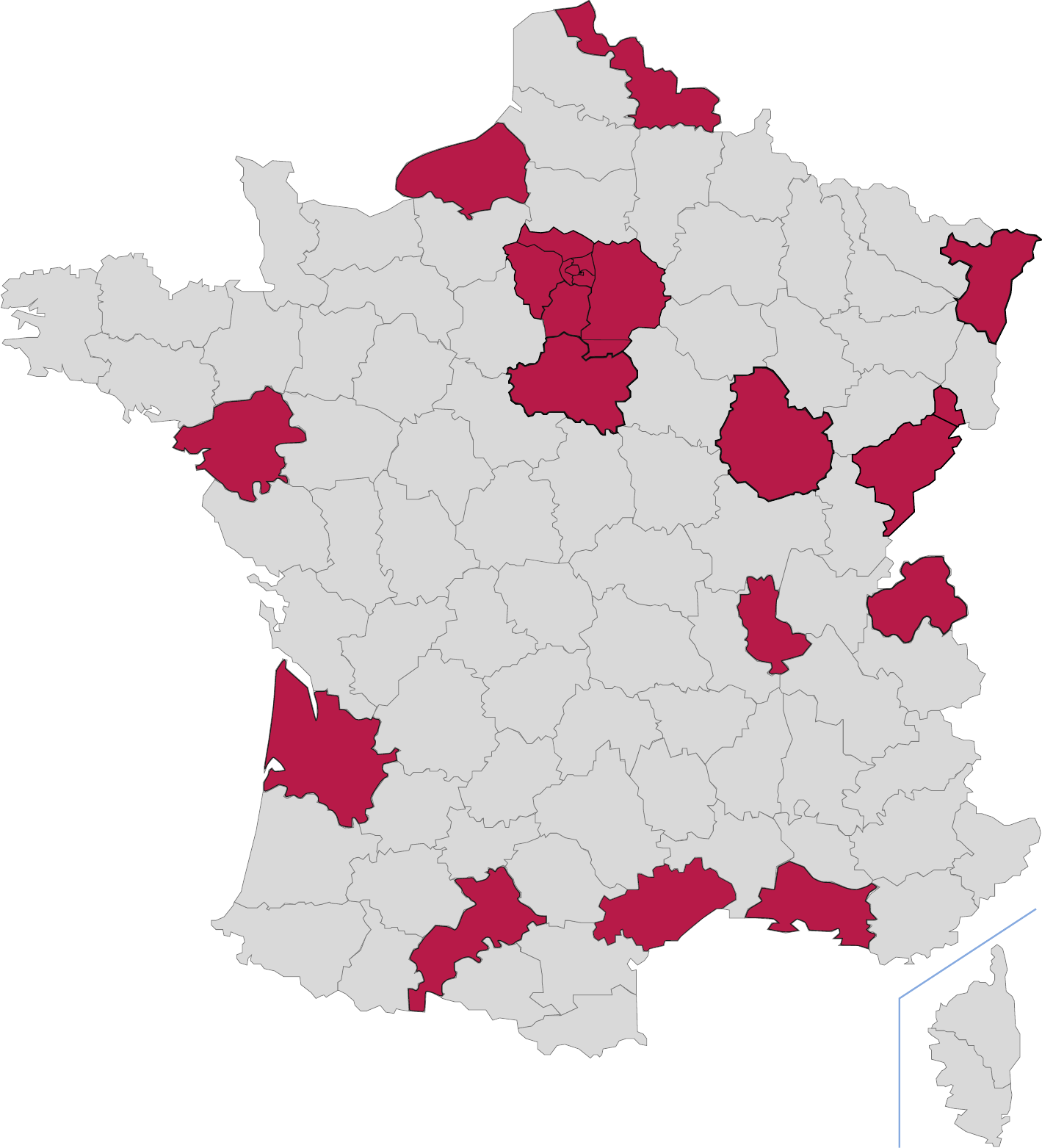} 
        \captionof{figure}{Selected administrative departments across the French territory.}
        \label{fig:map}
    \end{minipage}%
    \hfill 
    \begin{minipage}[c]{0.56\textwidth}
        \centering
        \scriptsize 
        \setlength{\tabcolsep}{4pt} 
        \begin{tabular}{lr | lr}
            \toprule
            \textbf{Class} & \textbf{Count} & \textbf{Class} & \textbf{Count} \\
            \midrule
            Building & 1.6M & Gold course & 8,040 \\
            Road & 294k & Marsh & 7,266 \\
            Vegetation & 263k & Stairway & 6,626 \\
            Forest & 215k & Airfield & 6,615 \\
            Path & 203k & Campground & 6,014 \\
            Woodland & 151k & Castle & 5,808 \\
            Gravel road & 136k & Military enclosure & 5,347 \\
            Watercourse & 132k & Water tank & 5,246 \\
            Reservoir & 77k & Shunting yard & 5,138 \\
            Bridge & 54k & Amusement park & 4,606 \\
            Vineyard & 49k & Tramway & 4,584 \\
            Indus. zone & 44k & Equastrian center & 4,468 \\
            Canal & 44k & Indoor sport & 4,356 \\
            Railroad & 39k & Hospital & 3,862 \\
            Orchard & 31k & Silo & 3,822 \\
            Bicycle path & 28k & Sport track & 3,574 \\
            Motorway & 27k & Power plant & 3,387 \\
            Greenhouse & 25k & Waste disposal & 3,365 \\
            L. sport field & 22k & Tower & 2,943 \\
            Public space & 22k & Chapel & 2,861 \\
            Water body & 31k & Pumping station & 2,747 \\
            Roundabout & 17k & Dam reservoir & 2,596 \\
            Parking lot & 16k & Runway & 2,266 \\
            Highway ramp & 15k & Grandstand & 2,211 \\
            Tennis court & 15k & Salt marsh & 2,093 \\
            Pond & 14k & Water tower & 2,078 \\
            Service track & 13k & Swimming pool & 1,980 \\
            Cemetery & 13k & Service area & 1,895 \\
            Factory & 11k & Sand & 1,894 \\
            S. sport field & 11k & Chairlift & 1,708 \\
            Stadium & 10k & Fire station & 1,689 \\
            Port & 8,990 & Fort & 1,534 \\
            Quarry & 8,831 & Livestock farm & 1,496 \\
            Church & 8,180 & Shooting range & 1,491 \\
            Water treat. plant & 8,077 & Train station & 1,487 \\
            \bottomrule
        \end{tabular}
        \captionof{table}{Top semantic classes (by occurrence).}
        \label{tab:classes}
    \end{minipage}
\end{figure}

\textbf{Geographical Coverage.} We selected 20 administrative departments (see Figure \ref{fig:map}) to encompass diverse environments (urban, alpine, maritime and rural), spanning $85,864 km^2$. 

\begin{figure}[hp!]
    \centering
    \begin{subfigure}[b]{1.0\textwidth}
        \centering
        \includegraphics[width=0.85\linewidth]{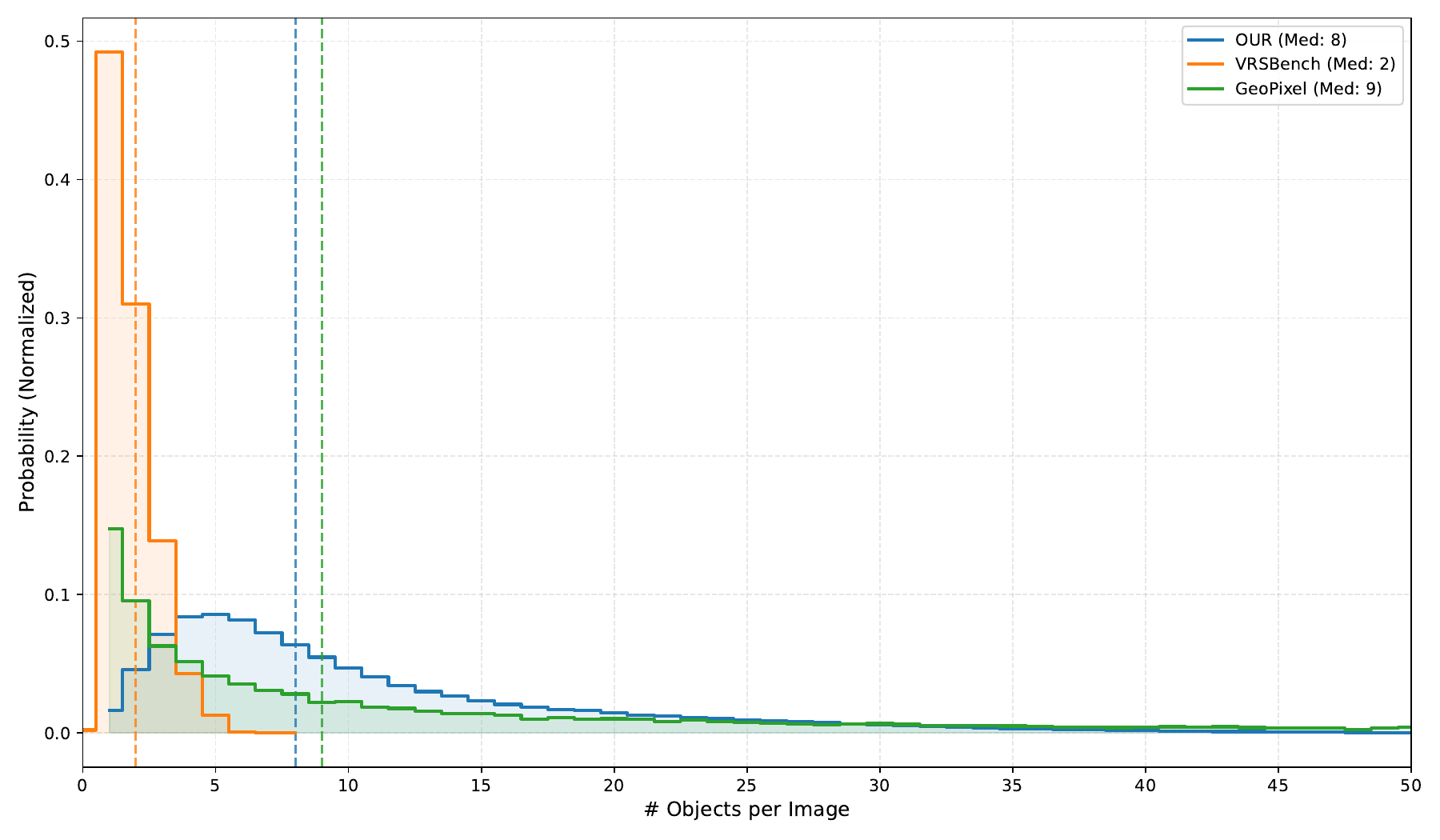}
        \caption{Objects per image distribution.}
        \label{fig:density_plot}
    \end{subfigure}
    
    \vspace{1em} 

    \begin{subfigure}[b]{1.0\textwidth}
        \centering
        \includegraphics[width=0.85\linewidth]{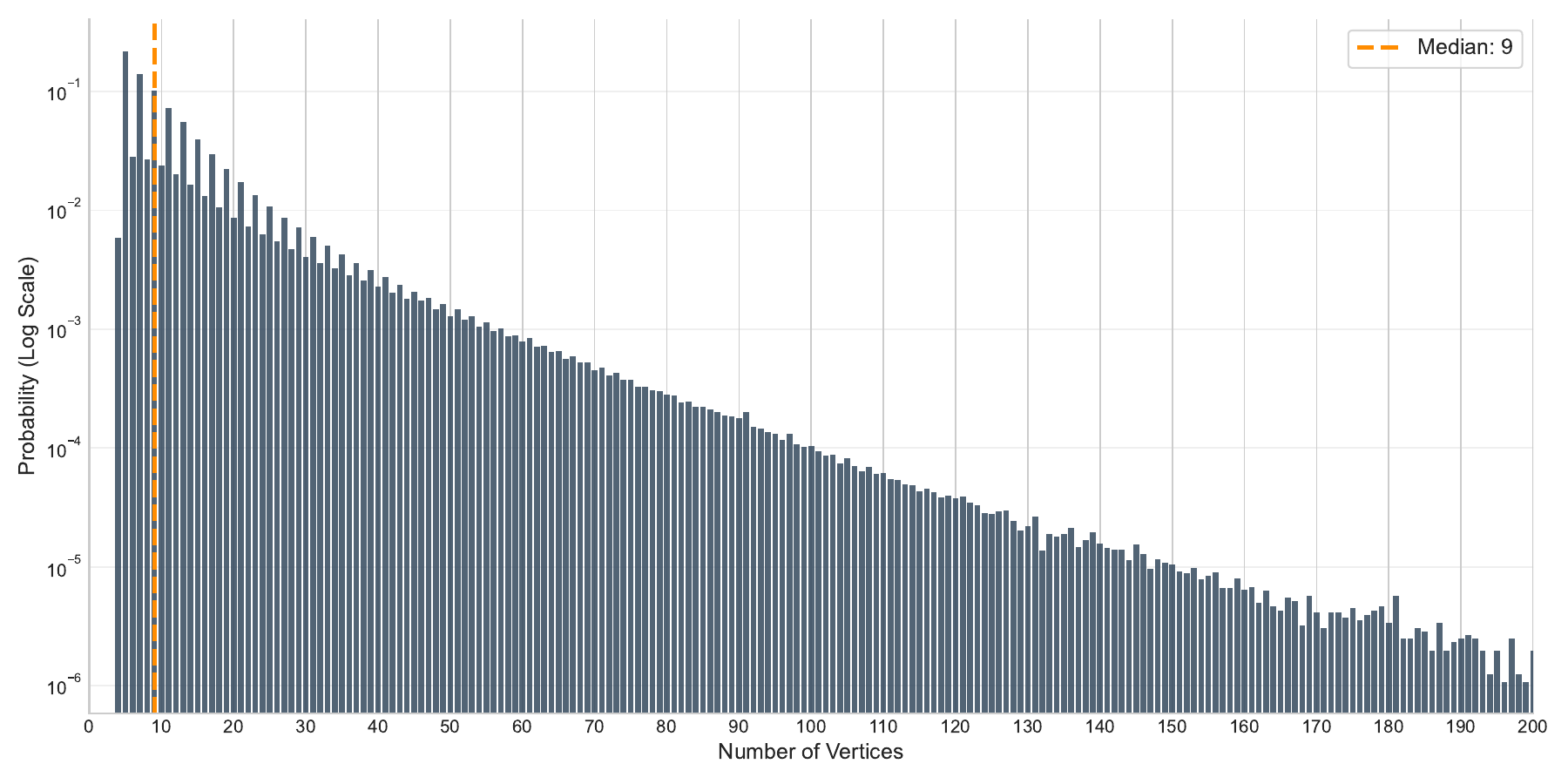}
        \caption{Vertex counts (Log Scale).}
        \label{fig:vertex_dist}
    \end{subfigure}
    
    \caption{Statistical analysis of the dataset showing a) the density of objects per image and b) the distribution of vertex counts across the geometric shapes.}
    \label{fig:combined_stats}
\end{figure}

\noindent
\textbf{Statistics.} 
As shown in Figure \ref{fig:density_plot}, our dataset maintains a high annotation density (median: 8 objects/image), outperforming standard benchmarks like VRSBench (median: 2). Although GeoPixel exhibits a marginally higher median (9), its density is heavily inflated by the inclusion of moving objects (e.g., vehicles), which can easily counts to over 400 objects per image. Moreover, we utilize native vector data instead of raster masks. Our annotations therefore consist of precise, human-verified geometric primitives that are inherently more efficient to process and topologically accurate (see Figure \ref{fig:vertex_dist}).

\noindent
\textbf{Data Source.} We utilize official data from IGN (French National Institute of Geographic and Forest Information), specifically \textit{BD ORTHO\textsuperscript{\textregistered}} (for imagery\footnote{\url{https://geoservices.ign.fr/bdortho}}) and \textit{BD TOPO\textsuperscript{\textregistered}} (vector data\footnote{\url{https://geoservices.ign.fr/bdtopo}}), released under Open Licence 2.0 (more details are available at \url{https://www.data.gouv.fr/datasets}).
Top semantic categories retrieved from the data are shown in Table \ref{tab:classes}.

\section{Implementation Details}
\paragraph{Architecture \& Training.} We use the LLaVA-1.6 architecture based on Vicuna 7B. The model is fine-tuned for 1 epoch on $8\times$ A100 (80GB) GPUs. We use LoRA (Low-Rank Adaptation) with rank $r=32$, alpha $\alpha=64$ and dropout $p=0.1$. We apply LoRA adapters to all linear layers of the language model, while the multi-modal projector is fully fine-tuned. The vision tower remains frozen during this stage. To ensure training efficiency, we also employ \textit{FlashAttention-2} and train in BFloat16 precision with TF32 enabled and DeepSpeed ZeRO-2.

\paragraph{Hyper-parameters.} We use the AdamW optimizer with a global batch size of 128. We employ a cosine learning rate scheduler with a warmup ratio of 0.03 and a peak learning rate of $2e^{-4}$. The weight decay is set to 0.0. The maximum sequence length is set to 4096 tokens.

\paragraph{Caption Generation.} The captioning workflow proceeds as follows: 
\begin{enumerate}
    \item We generate a structured raw caption using rule-based templates that list objects, group homogeneous elements and analyze spatial context (e.g., neighbors, inclusion, adjacency) via a $3 \times 3$ grid.
    \item The raw caption and image are fed into Gemma-3 27B, which refines the structured text into natural prose\footnote{We selected Gemma-3 after comparing state-of-the-art open models (DeepSeek R1, Qwen 3, Phi-4), where multimodal models demonstrated superior grounding performance.}.
\end{enumerate}

\begin{figure}[hp!]
    \centering
    \includegraphics[width=0.45\textwidth]{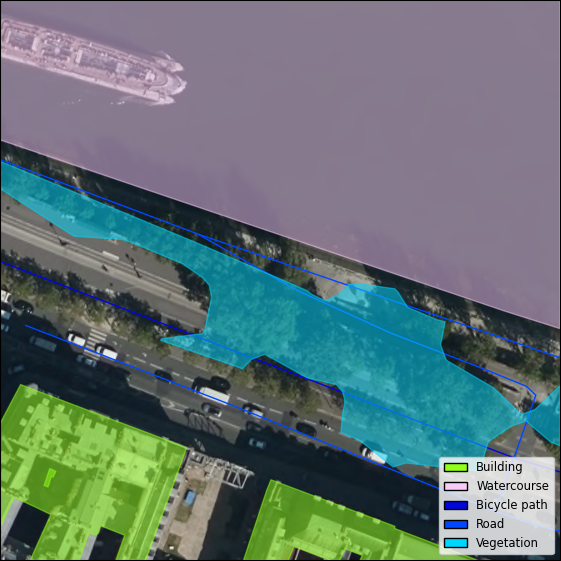}
    
    \vspace{1em}
    
    \begin{tcolorbox}[colback=gray!5!white, colframe=gray!50!black, arc=2pt, boxrule=0.5pt]
        \small
        \textbf{Raw caption:} \\
        The image contains: \\
        2 Buildings (Building2, Building1); 1 Watercourse (Watercourse1); 1 Path (Bicycle path1); 1 Road (Road1); 1 Vegetation (Vegetation1). \\
        More in detail: the Building2 spanning the bottom-left and bottom-center; the Watercourse1 is located on the top side; the Bicycle path1 is located on the center-left; the Road1 is located on the top-left; the Vegetation1 spanning the whole middle, top-left, bottom-center and bottom-right; the Building1 is located on the bottom-left. \\ 
        Building1 is partially included in Building2; Bicycle path1 is partially along Vegetation1. 
        \\
        Building2 is near: Vegetation1 on the right, Road1 on the bottom; Watercourse1 is near: Road on the top.
        
        \tcblower 
        
        \textbf{Final Caption:} \\
        This is an urban area with a mix of transportation infrastructure and green spaces alongside a watercourse. A large building dominates the lower-left portion of the scene, with a second building extending from and partially encompassing the first. Above these buildings, a road runs along the top-left, closely followed by a watercourse. A bicycle path curves through the center-left, running alongside a substantial area of vegetation that stretches across much of the image. The vegetation also borders the road, and extends along the base of the buildings.
    \end{tcolorbox}
    \caption{Example of a rule-based raw caption and the resulting synthetic caption refined by the VLM.}
    \label{fig:caption}
\end{figure}

We report in Figure \ref{fig:caption} the synthetic caption generated by re-elaborating the corresponding raw caption. Following this principle, we also generate synthetic referring expressions, yielding 197,399 referring expressions and 58,479 captions.

\noindent
\paragraph{Misalignment Identification.} For each object, we visually highlight its bounding box and crop the surrounding area to provide some context. We then use Gemini-2.5 Flash to identify possible misalignments (e.g., non-existing elements or hidden objects). Notably, the output of Gemini is only used to flag the data. We use the prompt reported in Figure \ref{fig:prompt}.

\begin{figure}[hbt!]
    \centering
    \begin{promptbox}[System Prompt: Visibility Filter]
    You are a Remote Sensing Data Quality Analyst. Your task is to perform a rapid visual inspection of the aerial image to determine if a feature is detectable within the red bounding box.
    
    \vspace{0.5em}
    \noindent \textbf{Core Goal: Filter Out Only ``Invisible'' Elements.} \\
    You are a visibility filter. You must keep any box where the object is partially visible, regardless of why it is cut off.
    
    \vspace{0.5em}
    \noindent \textbf{Criteria:}
    \begin{itemize}
        \item \textbf{OK:} The red box contains ANY discernible structure or feature.
        \begin{itemize}
            \item \textit{Partial Visibility:} Mark as OK if the object is partially under trees/shadows; cut off by the box edges or image borders (Truncated); or if only a small fragment of the object (e.g., a corner of a roof) is visible.
            \item \textit{Semantic Mismatch:} If the visible object matches the wrong class (e.g., `Garage' vs `House'), mark it OK.
        \end{itemize}
        \item \textbf{NOISE:} The feature is completely impossible to spot. This occurs ONLY if:
        \begin{enumerate}
            \item \textit{Full Occlusion:} The box is fully covered by tree canopy, dense shadow, or clouds.
            \item \textit{Empty:} The box clearly contains only empty ground, grass, or water with absolutely no structure present.
        \end{enumerate}
    \end{itemize}
    
    \vspace{0.5em}
    \noindent \textbf{Your Response Format:} \\
    Provide ONLY a single word from the two options below: \\
    \texttt{[OK | NOISE]}
    \end{promptbox}
    \caption{System prompt used by the VLM for the visibility filtering and misalignment identification task.}
    \label{fig:prompt}
\end{figure}

\clearpage
\section{Quantitative Results}
We provide here detailed benchmarks for Captioning, Classification, VQA in Table \ref{tab:group1} and Grounding tasks (i.e., Detection, Segmentation, etc.) in Table \ref{tab:group2}.

To address class imbalance and to ensure our evaluation is not disproportionately skewed by highly represented categories (detailed in Table \ref{tab:classes}), we provide a fine-grained analysis of model performance. Table \ref{tab:details} disaggregates the classification and object detection scores of the best-performing models across a diverse subset of semantic groups. This subset includes ubiquitous categories like \textit{Buildings} and \textit{Sport structures}, alongside highly specialized classes such as \textit{Heritage sites} (e.g., castles, archaeological remains), \textit{Civil infrastructure} (e.g., canals, aqueducts), \textit{Transport facilities} (e.g., parking lots, service areas) and \textit{Land use} zones.
The comparison reveals that existing baselines exhibit a severe performance drop on rare and specialized classes, relying heavily on common objects typically found in legacy datasets. Conversely, our model maintains robust performance across the entire semantic spectrum, confirming that its capabilities are not merely an artifact of overfitting to the most frequent and easily recognizable categories.

\begin{table}[!ht]
    \centering
    \begin{subtable}{.48\linewidth}
        \centering
        \scriptsize
        \begin{tabular}{l c c c}
            \toprule
            Model & BLEU@4 & METEOR & CIDEr \\
            \midrule
            Gemini-2.5 & 2.52 & 18.59 & 16.68 \\ \midrule
            LLaVA-1.5 & 1.05 & 17.13 & 18.74 \\
            LLaVA-1.6 & 1.00 & 18.32 & 17.93 \\ \midrule
            MiniGPT-v2 & 0.02 & 6.38 & 0.24 \\
            PaliGemma-2 & 1.07 & \underline{25.33} & 18.04 \\
            Ferret & 0.67 & 12.75 & 18.73 \\ \midrule
            GeoChat & 0.47 & 10.99 & 3.99 \\
            SkySenseGPT & 1.70 & 20.63 & \underline{24.03} \\
            VRSBench & 0.57 & 14.75 & 2.65 \\
            \midrule
            \textbf{GroundSet} & \textbf{18.54} & \textbf{40.96} & \textbf{104.48} \\
            \bottomrule
        \end{tabular}
        \caption{Captioning}
    \end{subtable}%
    \hfill
    \begin{subtable}{.48\linewidth}
        \centering
        \scriptsize
        \begin{tabular}{l c c c c}
            \toprule
            Model & P & R & F1 & acc \\
            \midrule
            Gemini-2.5 & \textbf{96.65} & 73.59 & 83.56 & 78.20 \\ \midrule
            LLaVA-1.5 & 91.73 & 82.75 & 87.01 & 80.89 \\
            LLaVA-1.6 & 92.04 & 81.34 & 86.36 & 80.13 \\ \midrule
            MiniGPT-v2 & 94.79 & 64.00 & 76.41 & 69.44 \\
            PaliGemma-2 & 92.93 & 78.90 & 85.34 & 79.04\\
            Ferret & 96.03 & 52.88 & 68.20 & 61.87 \\ \midrule
            GeoChat & 81.01 & 98.64 & 88.96 & 81.07 \\
            SkySenseGPT & 90.84 & 88.58 & \underline{89.69} & \underline{84.26} \\
            VRSBench & 77.33 & \textbf{100.00} & 87.22 & 77.33 \\
            \midrule
            \textbf{GroundSet} & \underline{96.19} & \underline{98.00} & \textbf{97.09} & \textbf{95.45} \\
            \bottomrule
        \end{tabular}
        \caption{VQA}
    \end{subtable}
    
    \vspace{0.4cm}
    
    \begin{subtable}{0.9\linewidth}
        \centering
        \scriptsize
        \begin{tabular}{l c c c c}
            \toprule
            Model & acc & acc@0.8 & acc@0.6 & acc@0.4 \\
            \midrule
            Gemini-2.5 & \underline{49.77} & \underline{49.84} & \underline{51.92} & \underline{58.57} \\ \midrule
            LLaVA-1.5 & 28.43 & 28.62 & 29.69 & 38.97 \\
            LLaVA-1.6 & 29.10 & 29.20 & 30.28 & 38.43 \\ \midrule
            MiniGPT-v2 & 6.87 & 6.97 & 7.52 & 14.11 \\
            PaliGemma-2 & - & - & - & - \\
            Ferret & 15.15 & 15.34 & 16.57 & 26.72 \\ \midrule
            GeoChat & 7.39 & 7.59 & 8.67 & 16.07 \\
            SkySenseGPT & 8.09 & 8.15 & 8.87 & 14.37 \\
            VRSBench & - & - & - & - \\ 
            \midrule
            \textbf{GroundSet} & \textbf{94.17} & \textbf{94.18} & \textbf{94.79} & \textbf{96.50} \\
            \bottomrule
        \end{tabular}
        \caption{Classification}
    \end{subtable}
    \caption{Best results in \textbf{bold}, second best \underline{underlined}.}
    \label{tab:group1}
\end{table}

\begin{table}[!ht]
    \centering
    \begin{subtable}{.49\linewidth}
        \centering
        \scriptsize
        \setlength{\tabcolsep}{3pt}
        \begin{tabular}{l c c c c c}
            \toprule
            Model & mIoU & P@0.5 & R@0.5 & F1@0.5 & acc@0.5 \\
            \midrule
            Gemini-2.5 & 8.37 & 3.40 & 4.30 & 3.79 & 3.40 \\ \midrule
            LLaVA-1.5 & 3.74 & 1.24 & 1.82 & 1.48 & 0.74 \\
            LLaVA-1.6 & 7.90 & 0.74 & 1.22 & 0.92 & 1.72 \\ \midrule
            MiniGPT-v2 & 20.89 & 19.33 & 8.38 & 11.69 & 6.21 \\
            PaliGemma-2 & \underline{34.70} & \underline{35.73} & 11.90 & 17.85 & \underline{12.33} \\
            Ferret & 28.02 & 26.89 & \underline{13.62} & \underline{18.08} & 9.95 \\ \midrule
            GeoChat & 14.59 & 7.80 & 4.27 & 5.52 & 2.84 \\
            SkySenseGPT & 18.63 & 7.04 & 2.01 & 3.13 & 4.65 \\
            VRSBench & 8.59 & 3.13 & 1.86 & 2.33 & 1.19 \\
            \midrule
            \textbf{GroundSet} & \textbf{42.04} & \textbf{46.14} & \textbf{53.31} & \textbf{49.47} & \textbf{34.37} \\
            \bottomrule
        \end{tabular}
        \caption{Object Detection}
    \end{subtable}%
    \hfill
    \begin{subtable}{.49\linewidth}
        \centering
        \scriptsize
        \setlength{\tabcolsep}{3pt}
        \begin{tabular}{l c c c c c}
            \toprule
            Model & mIoU & P@0.5 & R@0.5 & F1@0.5 & acc@0.5 \\
            \midrule
            Gemini-2.5 & 15.42 & 11.01 & 9.12 & 9.98 & 5.25 \\ \midrule
            LLaVA-1.5 & 6.07 & 4.74 & 2.15 & 2.96 & 1.50 \\
            LLaVA-1.6 & 8.86 & 4.77 & 3.39 & 3.96 & 2.02 \\ \midrule
            MiniGPT-v2 & 28.98 & 15.16 & 23.68 & 18.48 & 10.18 \\
            PaliGemma-2 & \underline{43.20} & 20.24 & 20.61 & 20.42 & 11.37 \\
            Ferret & 38.36 & \underline{33.01} & \underline{32.64} & \underline{32.83} & \underline{19.64} \\ \midrule
            GeoChat & 16.70 & 7.26 & 4.88 & 5.84 & 3.01 \\
            SkySenseGPT & 32.29 & 10.81 & 0.07 & 0.14 & 0.07 \\
            VRSBench & 20.88 & 12.07 & 5.19 & 7.26 & 3.77 \\
            \midrule
            \textbf{GroundSet} & \textbf{66.23} & \textbf{72.40} & \textbf{71.87} & \textbf{72.14} & \textbf{56.42} \\
            \bottomrule
        \end{tabular}
        \caption{Multi-target Detection}
    \end{subtable}
    
    \vspace{0.3cm}
    
    \begin{subtable}{.49\linewidth}
        \centering
        \scriptsize
        \setlength{\tabcolsep}{3pt}
        \begin{tabular}{l c c c c c}
            \toprule
            Model & mIoU & P@0.5 & R@0.5 & F1@0.5 & acc@0.5 \\
            \midrule
            Gemini-2.5 & 7.54 & 4.82 & 4.58 & 4.70 & 4.47 \\ \midrule
            LLaVA-1.5 & 4.49 & 2.64 & 2.79 & 2.71 & 1.37 \\
            LLaVA-1.6 & 6.33 & 3.97 & 4.24 & 4.10 & 2.97 \\ \midrule
            MiniGPT-v2 & \underline{17.37} &  \underline{17.01} &  \underline{17.96} & \underline{17.47} & \underline{9.57} \\
            PaliGemma-2 & - & - & - & - & - \\
            Ferret & 14.38 & 12.14 & 16.17 & 13.87 & 7.46 \\ \midrule
            GeoChat & 11.16 & 6.00 & 6.27 & 6.13 & 3.38 \\
            SkySenseGPT & 9.30 & 2.49 & 1.21 & 1.63 & 4.71 \\
            VRSBench & 6.27 & 3.31 & 3.20 & 3.25 & 1.67 \\
            \midrule
            \textbf{GroundSet} & \textbf{34.82} & \textbf{39.50} & \textbf{30.40} & \textbf{39.45} & \textbf{27.01} \\
            \bottomrule
        \end{tabular}
        \caption{REC}
    \end{subtable}%
    \hfill
    \begin{subtable}{.49\linewidth}
        \centering
        \scriptsize
        \setlength{\tabcolsep}{3pt}
        \begin{tabular}{l c c c c c}
            \toprule
            Model & mIoU & P@0.5 & R@0.5 & F1@0.5 & acc@0.5 \\
            \midrule
            Gemini-2.5 & 6.64 & 3.34 & 2.40 & 2.79 & 8.11 \\ \midrule
            LLaVA-1.5 & 1.33 & 0.14 & 0.36 & 0.20 & 1.97 \\
            LLaVA-1.6 & 1.78 & 0.15 & 0.05 & 0.07 & 9.69 \\ \midrule
            MiniGPT-v2 & - & - & - & - & - \\
            PaliGemma-2 & 19.72 & 15.06 & 15.19 & 15.12 & 8.50 \\
            Ferret & \underline{22.00} & \underline{19.31} & \underline{20.25} & \underline{19.77} & \underline{11.02} \\
            \midrule
            GeoChat & - & - & - & - & - \\
            SkySenseGPT & - & - & - & - & - \\
            VRSBench & - & - & - & - & - \\
            \midrule
            \textbf{GroundSet} & \textbf{41.00} & \textbf{44.11} & \textbf{45.21} & \textbf{44.65} & \textbf{33.63} \\
            \bottomrule
        \end{tabular}
        \caption{Segmentation}
    \end{subtable}
    \caption{Quantitative results for Grounding Tasks.}
    \label{tab:group2}
\end{table}

\begin{table}[htbp]
    \centering
    \renewcommand{\arraystretch}{1.3} 
    \setlength{\tabcolsep}{3pt}

    \begin{subtable}{\textwidth}
        \centering
        \begin{tabular}{lcccccc}
             & \textsf{Building} & \textsf{Heritage} & \textsf{Civil inf.} & \textsf{Sport} & \textsf{Transport} & \textsf{Land} \\ \hline
            \textsf{Gemini-2.5} & \textsf{66.45} & \textsf{63.89} & \textsf{24.26} & \textsf{50.94} & \textsf{43.52} & \textsf{53.85} \\
            \textsf{LLaVA-1.6} & \textsf{40.33} & \textsf{13.89} & \textsf{4.95} & \textsf{38.83} & \textsf{25.87} & \textsf{27.69} \\
            \textsf{Ferret} & \textsf{22.68} & \textsf{8.33} & \textsf{5.10} & \textsf{13.86} & \textsf{13.32} & \textsf{19.53} \\
            \textsf{SkySenseGPT} & \textsf{3.62} & \textsf{11.11} & \textsf{2.97} & \textsf{19.62} & \textsf{8.45} & \textsf{24.62} \\
            \hline
            \textsf{GroundSet} & \textsf{97.45} & \textsf{88.89} & \textsf{81.68} & \textsf{96.03} & \textsf{93.23} & \textsf{90.77} \\
        \end{tabular}
        \caption{Acc@0.8 score for the classification task}
        \label{tab:det_class}
    \end{subtable}

    \vspace{2.5em} 

    \begin{subtable}{\textwidth}
        \centering
        \begin{tabular}{lccccc}
             & \textsf{Building} & \textsf{Heritage} & \textsf{Civil inf.} & \textsf{Sport} & \textsf{Transport}  \\ \hline
            \textsf{MiniGPT-v2} & \textsf{11.31} & \textsf{1.75} & \textsf{0.98} & \textsf{25.59} & \textsf{5.79} \\
            \textsf{PaliGemma-2} & \textsf{15.15} & \textsf{0.00} & \textsf{0.00} & \textsf{39.64} & \textsf{17.40} \\
            \textsf{Ferret} & \textsf{16.46} & \textsf{0.00} & \textsf{6.54} & \textsf{36.09} & \textsf{15.36} \\
            \textsf{GeoChat} & \textsf{5.47} & \textsf{0.00} & \textsf{0.00} & \textsf{13.42} & \textsf{1.10} \\
            \hline
            \textsf{GroundSet} & \textsf{46.97} & \textsf{33.33} & \textsf{48.57} & \textsf{83.92} & \textsf{45.50} \\
        \end{tabular}
        \caption{F1@0.5 score for the detection task}
        \label{tab:det_det}
    \end{subtable}
    \caption{Detailed results of best performing models for classification (\ref{tab:det_class}) and detection (\ref{tab:det_det}) on different semantic categories.}
    \label{tab:details}
\end{table}

\section{Qualitative Results}
\begin{figure}[p]
    \centering
    \includegraphics[width=0.95\textwidth]{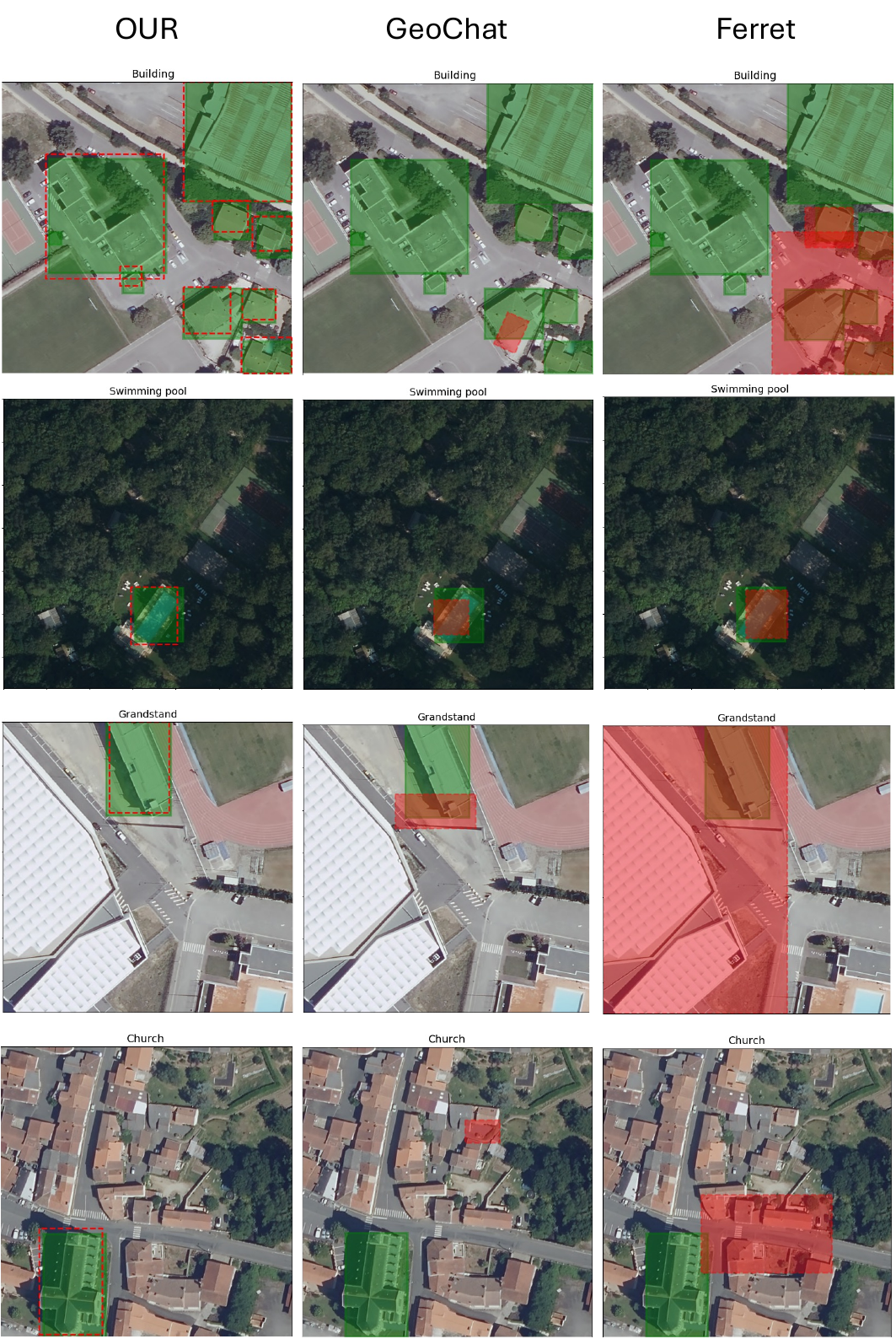}
    \caption{Qualitative comparisons on \textbf{Object Detection}. \textcolor{green}{Green} regions represent Ground Truth annotations, while \textcolor{red}{Red} regions indicate model predictions. Our model (left) shows superior overlap compared to GeoChat and Ferret.}
    \label{fig:ex_det}
\end{figure}

In Figure \ref{fig:ex_det}, we present qualitative comparisons on the object detection task, demonstrating that our model achieves superior overlap with the ground truth compared to GeoChat and Ferret. As observed in the \textit{Grandstand} and \textit{Church} examples, the baseline models frequently over-predict the spatial extent of the objects, resulting in massive, imprecise bounding boxes. 

\begin{figure}[p]
    \centering
    \includegraphics[width=0.95\textwidth]{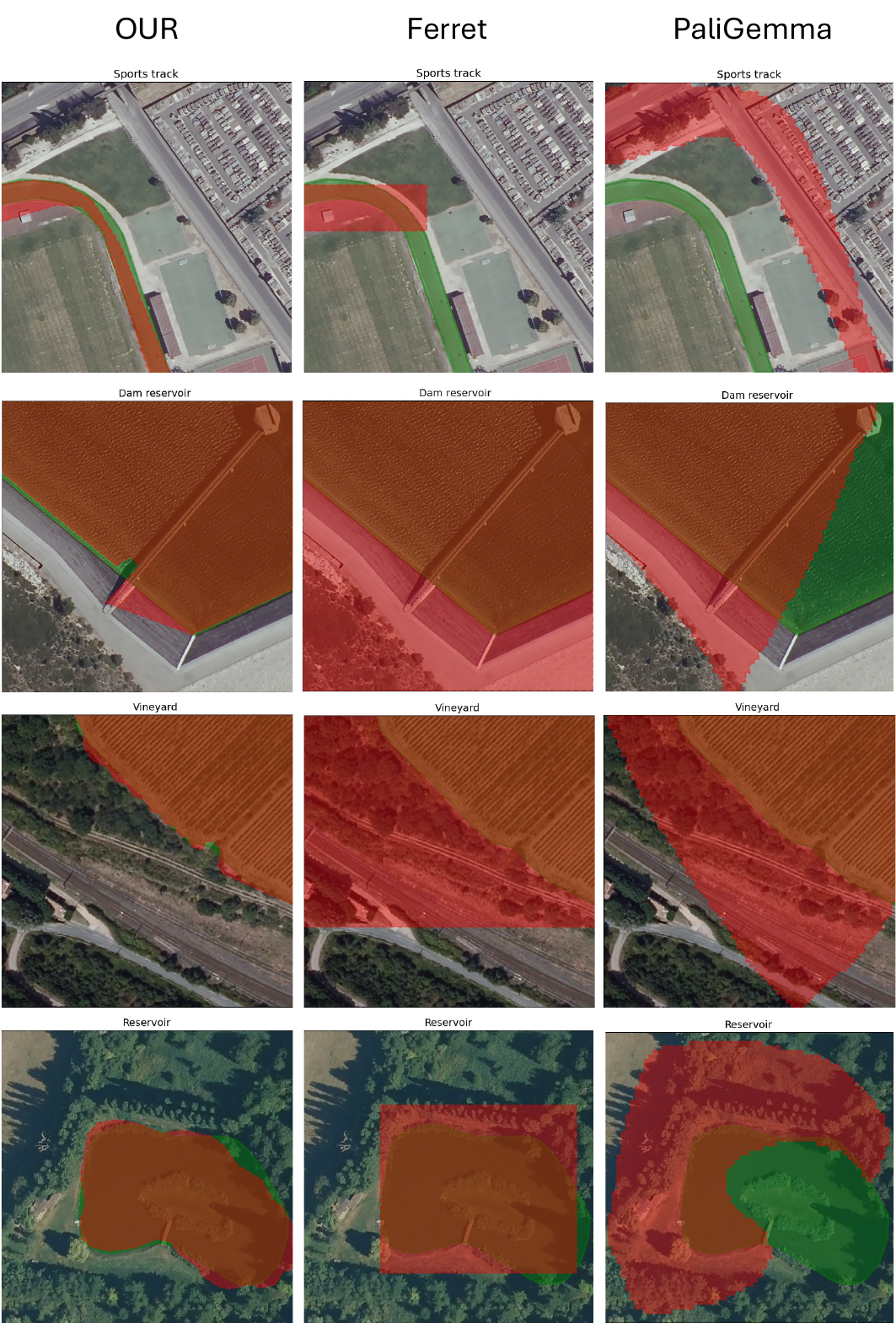}
    \caption{Qualitative comparisons on \textbf{Segmentation}. \textcolor{green}{Green} represents the Ground Truth and \textcolor{red}{Red} the model predictions. The proposed method handles complex boundaries (e.g., vineyards, reservoirs) more accurately than baselines.}
    \label{fig:ex_seg}
\end{figure}

Figure \ref{fig:ex_seg} illustrates our model's capabilities in the segmentation task when compared against Ferret and PaliGemma. Standard baselines consistently struggle to accurately map irregular boundaries, often smearing predictions over vast areas, as seen in the Vineyard and Dam reservoir samples.

\begin{figure}[p]
    \centering
    \includegraphics[width=0.95\textwidth]{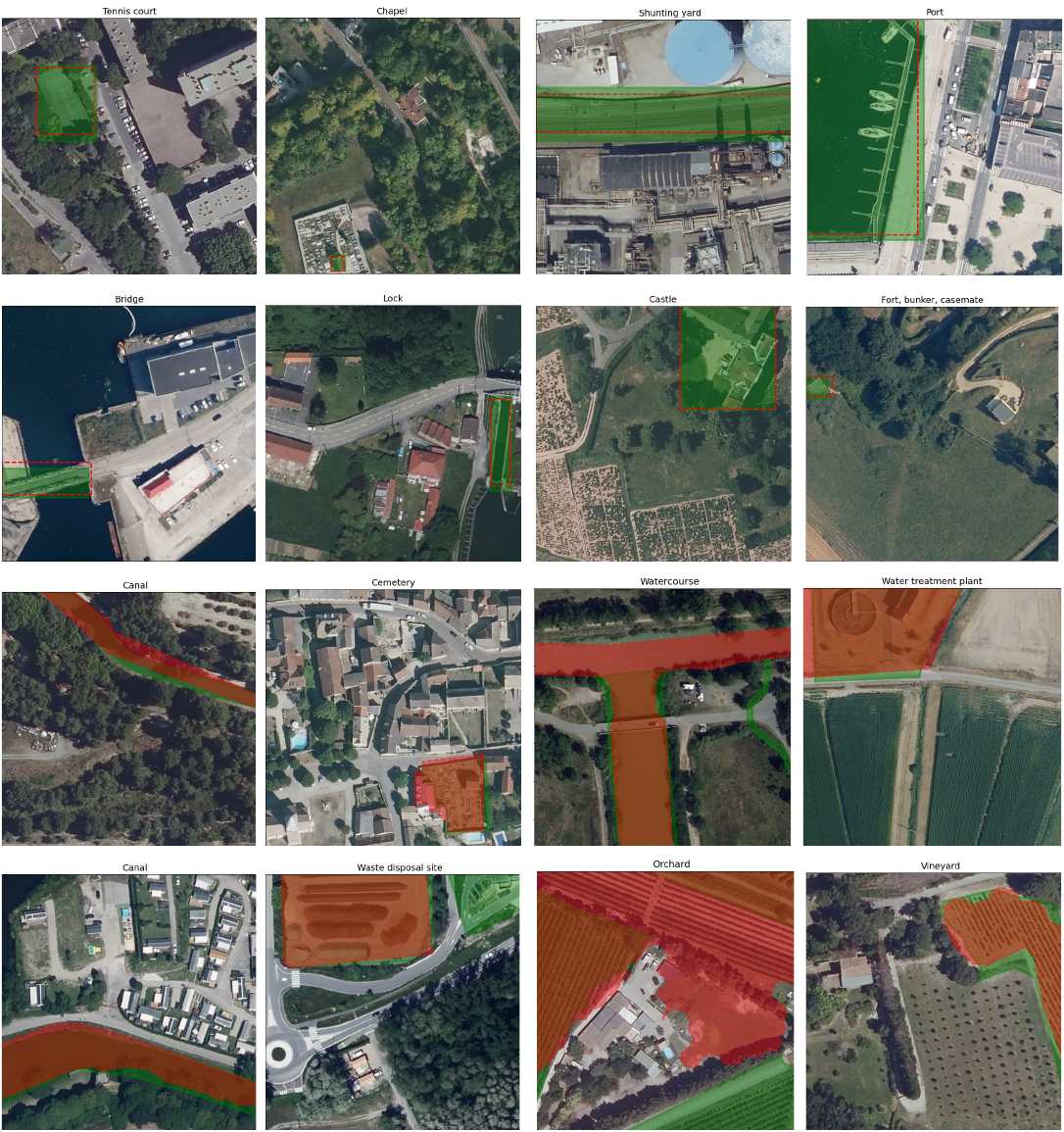}
    \caption{Detection and segmentation performance of our proposed baseline on a wide range of new semantic categories. \textcolor{green}{Green} represents the Ground Truth, \textcolor{red}{Red} the model predictions.}
    \label{fig:ex_pred}
\end{figure}

Furthermore, to visually substantiate our quantitative findings, Figure \ref{fig:ex_pred} showcases our baseline's predictions on a wide range of new and specialized semantic categories. 
\FloatBarrier

\end{document}